\pdfoutput=1

\documentclass[11pt]{article}

\usepackage[dvipsnames]{xcolor}
\usepackage[preprint]{acl}

\usepackage{times}
\usepackage{latexsym}

\usepackage[T1]{fontenc}

\usepackage[utf8]{inputenc}

\usepackage{microtype}

\usepackage{inconsolata}

\usepackage{graphicx}
\usepackage{amsmath}
\usepackage{todonotes}
\usepackage{booktabs}
\usepackage{multirow}

\title{DiCoRe: Enhancing Zero-shot Event Detection via \\ Divergent-Convergent LLM Reasoning}

\author{Tanmay Parekh$^\dagger$ \ \ \ \ \
Kartik Mehta$^\ddagger$ \ \ \ \ \
Ninareh Mehrabi$^\mathsection$\thanks{Work done while at Amazon.} \\
{\bf Kai-Wei Chang$^\dagger$ \ \ \ \ \
Nanyun Peng$^\dagger$} \\
$^\dagger$Computer Science Department, University of California, Los Angeles \\
$^\ddagger$Amazon AGI Foundations \quad \quad \quad \quad $^\mathsection$Meta\\
\texttt{\{tparekh, kwchang, violetpeng\}@cs.ucla.edu} \\
  }

\begin{document}
\maketitle

\newcommand{\modelName}{\textsc{DiCoRe}}

\newcommand{\redtext}[1]{\textcolor{red}{\textbf{#1}}}
\newcommand{\greentext}[1]{\textcolor{OliveGreen}{\textbf{#1}}}
\newcommand{\bluetext}[1]{\textcolor{blue}{#1}}
\newcommand{\browntext}[1]{\textcolor{brown}{#1}}

\newcommand\blfootnote[1]{%
  \begingroup
  \renewcommand\thefootnote{}\footnote{#1}%
  \addtocounter{footnote}{-1}%
  \endgroup
}

\newcommand{\mypar}[1]{\vspace{0.35em}\noindent\textbf{#1}}
\newcommand{\SideNote}[2]{\todo[color=#1,size=\small]{#2}} 

\newcommand{\tanmay}[1]{\SideNote{orange!40}{#1 --tanmay}}

\newcommand{\nina}[1]{\SideNote{green!40}{#1 --nina}}

\begin{abstract}

Zero-shot Event Detection (ED), the task of identifying event mentions in natural language text without any training data, is critical for document understanding in specialized domains.
Understanding the complex event ontology, extracting domain-specific triggers from the passage, and structuring them appropriately overloads and limits the utility of Large Language Models (LLMs) for zero-shot ED.
To this end, we propose \modelName{}, a divergent-convergent reasoning framework that decouples the task of ED using Dreamer and Grounder.
Dreamer encourages divergent reasoning through open-ended event discovery, which helps to boost event coverage.
Conversely, Grounder introduces convergent reasoning to align the free-form predictions with the task-specific instructions using finite-state machine guided constrained decoding.
Additionally, an LLM-Judge verifies the final outputs to ensure high precision.
Through extensive experiments on six datasets across five domains and nine LLMs, we demonstrate how \modelName{} consistently outperforms prior zero-shot, transfer-learning, and reasoning baselines, achieving 4–7\% average F1 gains over the best baseline -- establishing \modelName{} as a strong zero-shot ED framework.


\end{abstract}

\section{Introduction}
\label{sec:introduction}

\begin{figure}[t]
    \centering
    \includegraphics[width=\linewidth]{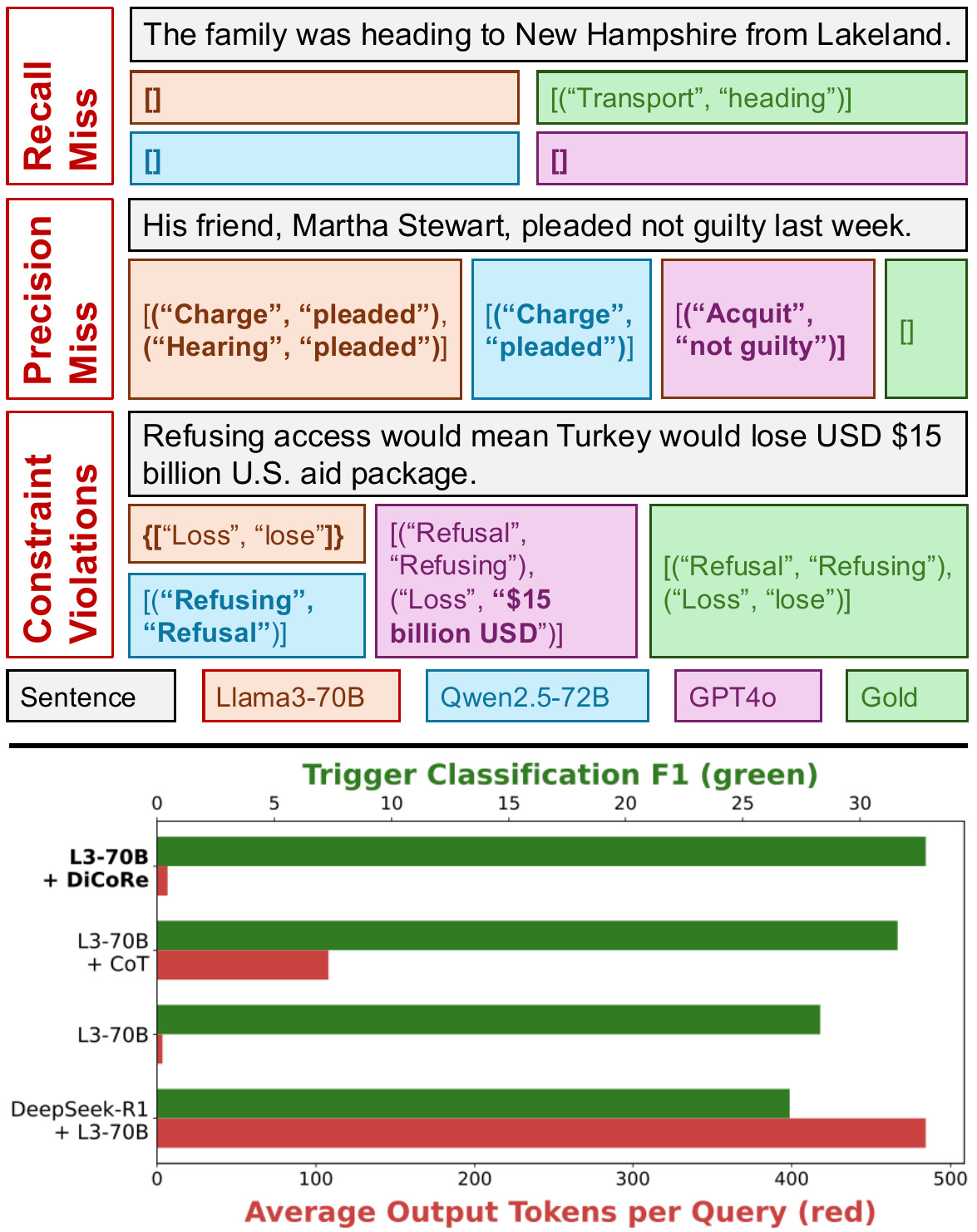}
    \caption{(top) Illustration of how prompting LLMs directly for Event Detection (ED) with all the task constraints can lead to precision, recall, and constraint violations (incorrect JSON, trigger not in sentence) across various LLMs. The errors are highlighted in \textbf{bold}. (bottom) Highlighting the superior model performance (\greentext{green bars}) of our proposed \modelName{} with minimal inference cost (\redtext{red bars}) relative to reasoning baselines.}
    \label{fig:teaser}
\end{figure}

Event Detection (ED) is the task of identifying events by extracting and labeling event triggers \cite{sundheim-1992-overview, doddington-etal-2004-automatic}.
ED aids in various downstream applications, including news monitoring \cite{news-monitoring}, biomedical literature mining \cite{mlee}, epidemic forecasting \cite{parekh-etal-2024-speed, parekh-etal-2024-event}, and legal understanding \cite{legal-application}.
Training effective ED models requires large amounts of expert-annotated domain-specific data, which is highly costly and labor-intensive.
This underlines the need to develop zero-shot systems that can perform ED robustly without using any training data.

Recently, large language models (LLMs) have shown strong zero-shot performance across various tasks \cite{instruct-gpt, llm-survey}.
However, their effectiveness on ED remains limited \cite{chatgpt-ed, huang-etal-2024-textee}, due to the requirement of extensive domain knowledge and the complex structural nature of ED.
ED requires deep reasoning and imposes several intertwined constraints: study of the large, closed event ontology and ensuring the event types must be chosen from it; semantic understanding of the input passage and precisely identifying domain-specific triggers within it; and conforming the output to a strict, machine-parsable structured format.
Encoding these constraints as natural language instructions in the prompt overloads the LLM cognitively, making it harder to effectively apply its reasoning skills \cite{tam-etal-2024-speak}.
This increased difficulty in reasoning causes failures, such as missing relevant events, predicting irrelevant ones, and struggling to follow the expected format, as shown in Figure~\ref{fig:teaser}.


To this end, we propose \modelName, a novel pipeline introducing \textbf{Di}vergent-\textbf{Co}nvergent \textbf{Re}asoning, that facilitates better ED performance by reducing the cognitive burden of constraint adherence on the LLM.
\modelName{} comprises two major components in a pipeline: Dreamer and Grounder.
(1) Dreamer fosters divergent reasoning by prompting in an unconstrained, open-ended manner.
This encourages broad semantic exploration of potential event mentions by removing rigid task constraints and, in turn, boosts the recall.
(2) Grounder introduces convergent reasoning by mapping Dreamer’s free-form predictions to the task-specific closed event ontology.
To alleviate the constraint adherence burden on the LLM, we employ a finite-state machine (FSM) to encode structural and task-specific constraints. 
This FSM guides the generation process through constrained decoding, ensuring that the output adheres to the task requirements.
Finally, we add an LLM-Judge to verify the grounded predictions against the original task instructions, ensuring high precision by filtering irrelevant predictions.

We conduct extensive experiments on six datasets from five domains across nine LLMs.
Compared with various existing LLM inference works \cite{chatgpt-ed, wang-etal-2023-code4struct, fig}, we show how \modelName{} performs the best with average improvements of 4-5\% F1 Trigger Classification and 5.5-6.5\% F1 Event Identification over the best baselines.
\modelName, without any training, also consistently improves over transfer-learning baselines \cite{hsu-etal-2022-degree, gollie} fine-tuned on 15-30k datapoints by at least 5-12\% F1.
Furthermore, we demonstrate that \modelName{} provides 1-2\% F1 gains while using 15-55x fewer inference tokens relative to strong thinking-based models and chain-of-thought (CoT), highlighting the significance of our proposed divergent-convergent reasoning.


In summary, we make the following contributions:
(1) We propose Dreamer, introducing divergent reasoning to improve event coverage.
(2) We develop Grounder, performing convergent reasoning to align free-form predictions to the event ontology.
(3) We design FSM-guided decoding to enforce task-specific structure during inference.
Through extensive evaluations across six datasets, five domains, and nine LLMs, we demonstrate the generalizability and efficacy of \modelName, establishing it as a robust zero-shot ED framework.
We will release the code at \url{https://github.com/PlusLabNLP/DiCoRe}.

\begin{figure}
    \centering
    \includegraphics[width=\linewidth]{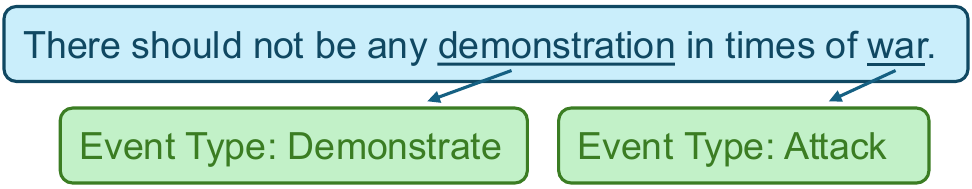}
    \caption{Illustration example for the task of Event Detection. Here, the blue box is the input sentence, and the green boxes are the event mentions. The underlined words indicate the event triggers.}
    \label{fig:ed-example}
\end{figure}

\section{Background and Related Works}
\label{sec:background}

Event Detection (ED) is the task of identifying event mentions from input text/document $X$ based on a pre-defined ontology \cite{sundheim-1992-overview, grishman-sundheim-1996-message, doddington-etal-2004-automatic}.
We follow previous works \cite{doddington-etal-2004-automatic} to define an \textit{event} as something that happens or describes a change of state.
Each event is labeled by an \textit{event type} $e$ and the list of pre-defined event types constitutes an \textit{event ontology} $\mathcal{E}$.
An \textit{event trigger} $t$ is defined as the most distinctive word/phrase that indicates the event's presence in the text $X$.
The trigger-event type pair $(t,e)$ is jointly referred to as the \textit{event mention}.
The extraction of trigger words from the text and classifying them into one or more event types from the event ontology is the task of Event Detection, described by $f$ below.
\begin{align}
    [(e_1,t_1), ... (e_n, t_n)] = f(X; \mathcal{E}) \nonumber
\end{align}
We provide an illustration of the task in Figure~\ref{fig:ed-example}, wherein \textit{demonstration} and \textit{war} indicate the mentions of \textit{Demonstrate} and \textit{Attack} events, respectively, in the sentence.

\paragraph{Event Detection:}
Traditionally, ACE05 \cite{doddington-etal-2004-automatic} and ERE \cite{song-etal-2015-light} have been traditionally utilized for developing various sequence-tagging \cite{wadden-etal-2019-entity, hsu-etal-2023-tagprime} and generative \cite{li-etal-2021-document, hsu-etal-2023-ampere} models.
However, procuring expert-annotated data in specialized domains like biomedicine, law, cybersecurity, etc. is an expensive and labor-intensive task, leading to explorations in zero-shot and low-resource ED.

\paragraph{Zero-shot Event Detection:}
Recently, various diverse datasets such as MAVEN \cite{wang-etal-2020-maven}, FewEvent \cite{fewevent}, GENEVA \cite{parekh-etal-2023-geneva} in general domain, GENIA2011 \cite{kim-etal-2011-overview-genia}, GENIA2013 \cite{kim-etal-2013-genia} in biomedical, CASIE \cite{casie} in cybersecurity, PHEE \cite{sun-etal-2022-phee} in pharmacovigilance, SPEED \cite{parekh-etal-2024-event}, SPEED++ \cite{parekh-etal-2024-speed} in epidemiology, etc. have been developed.
To explore generalizability across these domains/datasets, initial works posed ED as a question-answering \cite{du-cardie-2020-event} or machine-reading comprehension problem \cite{liu-etal-2020-event}.
Various works explored transfer and joint learning across various IE tasks to build more universal IE models \cite{lu-etal-2022-unified, lasuie, li-etal-2024-knowcoder}.
Some works have explored posing ED as a generative text-to-text approach with event-based templates \cite{lu-etal-2021-text2event, li-etal-2021-document, hsu-etal-2022-degree}, even for zero-shot cross-lingual transfer \cite{huang-etal-2022-multilingual-generative, parekh-etal-2024-contextual}.
However, these works require source data training for zero-shot transfer, limiting their utility.
Recent works have also explored the utility of zero-shot prompting with LLMs - concluding their sub-par performance \cite{chatgpt-ed, ie-llm-eval}.
Other works have explored utilizing LLMs for data generation \cite{star, zhang-etal-2024-unexpected, fig} to aid better generalizability.
In our work, we focus on improving LLMs' zero-shot task generalizability to ED without any model fine-tuning.

\paragraph{Unconstraining LLMs for Better Reasoning:}
LLMs show immense language understanding and generation capabilities, but they need instructions and constraints to aid in meaningful human tasks \cite{instruction-tuning}.
However, imposing constraints also reduces LLM reasoning capabilities \cite{tam-etal-2024-speak, tian-etal-2024-macgyver, crane}.
To this end, works have explored constrained decoding by altering the output probability distribution \cite{outlines, grammar-masking, ctrl-g}.
Some works explore grammar-aligned decoding strategies \cite{geng-etal-2023-grammar, grammar-aligned-decoding}.
However, such strict enforcement has been shown to hurt LLMs' generations.
Recently, \citet{tam-etal-2024-speak} explored better prompt design on math reasoning to unburden the constraints on the LLM.
With similar inspiration, we explore decoupling LLMs from constraints to improve reasoning in our work.
Although, we only explore the task of Event Detection, we believe our work could benefit other structured tasks in Information Extraction \cite{li-etal-2023-revisiting-large, wang-etal-2025-gpt}, Document Understanding \cite{suvarna-etal-2024-qudselect}, Question Answering \cite{rajpurkar-etal-2016-squad, parekh-etal-2025-dynamic}, and Dialogue Generation \cite{parekh-etal-2020-understanding, tartan}.


\begin{figure*}[t]
    \centering
    \includegraphics[width=\linewidth]{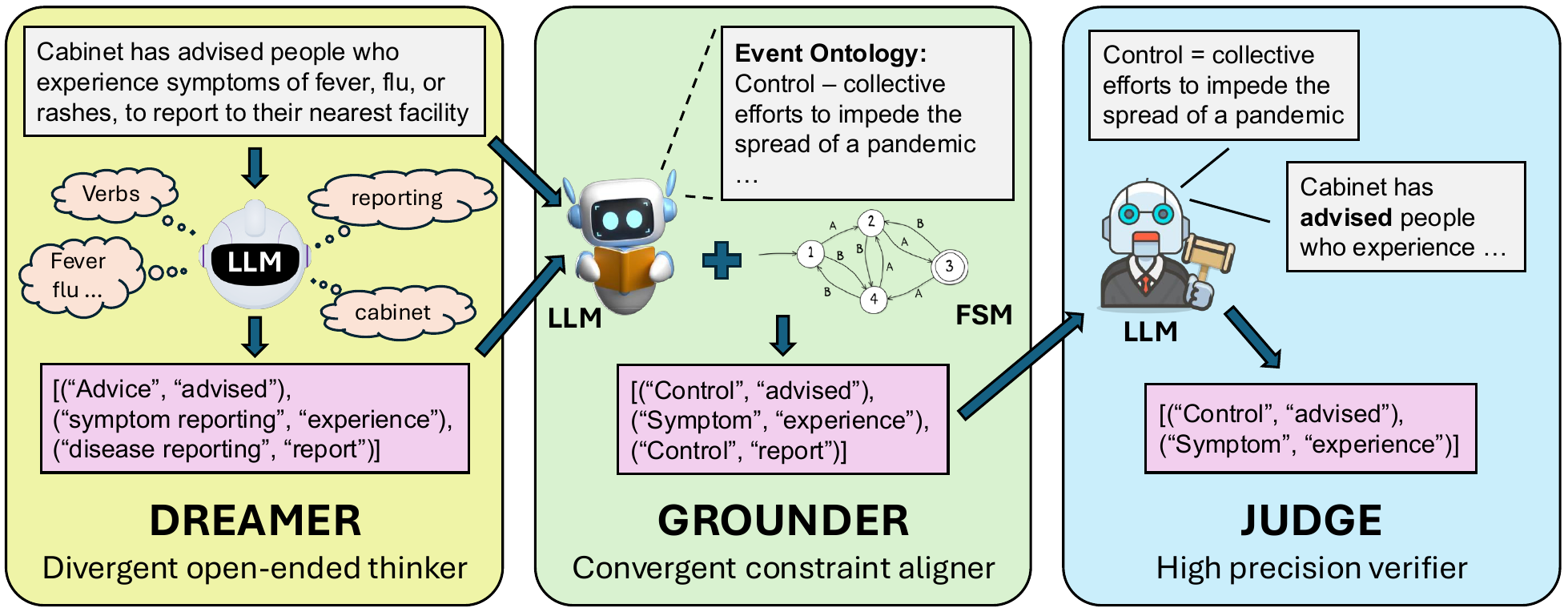}
    \caption{Illustration of our \modelName{} pipeline. First, the Dreamer reasons divergently in an open-ended unconstrained manner about all potential events in the text and generates free-form event names. Next, the Grounder reads the event ontology and convergently grounds the free-form predictions to one of the event types. It uses finite-state machine (FSM) guided constrained decoding to enforce task-specific constraints. Finally, the Judge evaluates each prediction and verifies its validity at a holistic scale.}
    \label{fig:methodology}
\end{figure*}


\section{Methodology}
\label{sec:methodology}

In our work, we frame ED through a generative outlook $f_{gen}$ \cite{tanl, huang-etal-2022-multilingual-generative} as they provide stronger zero-shot performance \cite{hsu-etal-2022-degree} and are better suited for LLMs.
We consider a structured list of tuples as the output generation as they provide stronger performance (\S~\ref{sec:appendix-structured-output-analysis}) and are easy to parse \cite{wang-etal-2023-code4struct}. 
However, these considerations introduce constraints (encoded as task instructions in LLM prompt) like the predicted event is from the provided list, the predicted trigger phrase is in the input text, and the output generation follows the JSON format, as technically described below.
\begin{align}
    Y &= f_{gen}(X;\mathcal{E}) \quad \quad \text{where} \nonumber \\
    Y &= ``[ (e_1, t_1), ... (e_n, t_n) ]" \label{eq:json-constraint} \\
    t &\in X \quad \quad \forall t \in \{t_1, ... t_n\} \label{eq:trig-constraint} \\
    e &\in \mathcal{E} \quad \quad \forall e \in \{e_1, ... e_n\} \label{eq:event-constraint}
\end{align}

We argue that these structured constraints inherent to ED (Eq.~\ref{eq:json-constraint}-\ref{eq:event-constraint}) increase the cognitive load on LLMs, making reasoning more difficult \cite{tam-etal-2024-speak}.
This is one of the contributing factors to LLMs' subpar performance for ED \cite{huang-etal-2024-textee}.
To address this, we propose \modelName, a novel pipeline that decouples and reduces constraint adherence through divergent open-ended discovery, convergent alignment, and constrained decoding.
\modelName{} is lightweight, does not require additional training, and can be seamlessly applied to any LLM.
Specifically, \modelName{} comprises a three-stage pipeline of a Dreamer-Grounder-Judge, as illustrated in Figure~\ref{fig:methodology}, and described below.

\subsection{Dreamer}
\label{sec:dreamer}

Our first component, Dreamer \textit{aka Divergent open-ended thinker}, is designed to promote open-ended divergent discovery, encouraging the LLM to achieve high recall by freely identifying potential events without being constrained by the predefined event ontology.
Specifically, the Dreamer component $f_d$ removes the task-specific event constraint (Eq.~\ref{eq:event-constraint}), relaxes the trigger constraint (Eq.~\ref{eq:trig-constraint}), and prompts the LLM to extract event mentions directly from the input sentence $X$ as
$$
    Y_{d} = ``[(e'_1,t_1), ... (e'_n,t_n)]"= f_{d}(X)
$$
where each $e'_i$ is a free-form LLM-generated natural language event name. Notably, $e'_i$ need not adhere to the event ontology $\mathcal{E}$.
We provide an illustration of the LLM prompt in Figure~\ref{fig:dreamer-prompt}.

By removing explicit references to the event ontology, the instructions become less restrictive and more semantically intuitive for the LLM.
This simplification enables the model to divergently reason on the underlying semantics of the text, rather than rigidly aligning with predefined categories.
This open-ended setup encourages broader event discovery, improving recall by allowing the model to identify diverse or implicit event types.
Though it may lower precision, it produces a rich candidate set for downstream refinement.

\subsection{Grounder}
\label{sec:grounder}

The second component, Grounder \textit{aka Convergent constraint aligner}, convergently aligns the Dreamer’s open-ended predictions $Y_d$ with the closed, task-specific event ontology $\mathcal{E}$, while filtering the events that are not mappable.
Technically, the Grounder component $f_g$ infuses the original task-specific constraints into the prompt to generate the grounded event mentions $Y_g$ as
\begin{align}
    Y_g = ``[(e_1,t_1), ... (e_m,t_m)]"= f_{g}(X;\mathcal{E},Y_d) \nonumber
\end{align}
An illustration of the Grounder prompt and expected output is shown in Figure~\ref{fig:grounder-prompt}.

\begin{figure}[t]
    \centering
    \includegraphics[width=\linewidth]{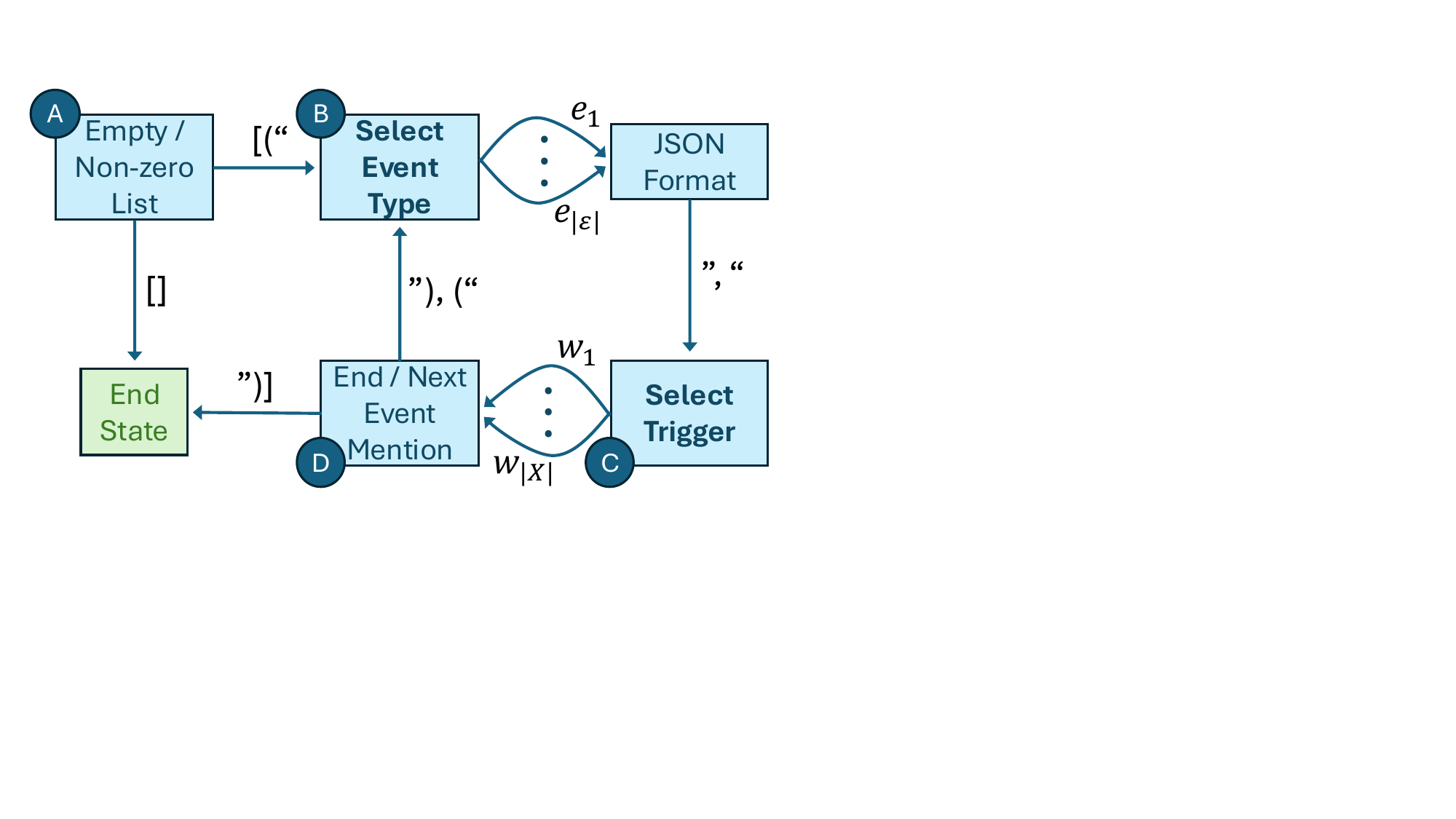}
    \caption{Finite state machine (FSM) illustration for guiding decoding to enforce constraints. Here $e_1,\dots,e_{|\mathcal{E}|} \in \mathcal{E}$ represent all the possible event types and $w_1,\dots,w_{|X|} \in X$ represent the atomized phrases in the sentence $X$.}
    \label{fig:grounder-fsm}
\end{figure}

\begin{table*}[t!]
\centering
\setlength{\tabcolsep}{3.5pt}
\resizebox{\textwidth}{!}{
\begin{tabular}{ll|ccc|ccc|ccc|ccc|ccc|ccc|ccc}
    \toprule
    \multirow{2}{*}{\textbf{LLM}} & \textbf{Prompt} & \multicolumn{3}{c|}{\textbf{MAVEN} (168)} & \multicolumn{3}{c|}{\textbf{FewEvent} (100)} & \multicolumn{3}{c|}{\textbf{ACE} (33)} & \multicolumn{3}{c|}{\textbf{GENIA} (9)} & \multicolumn{3}{c|}{\textbf{SPEED} (7)} & \multicolumn{3}{c|}{\textbf{CASIE} (5)} & \multicolumn{3}{c}{\textbf{Average}} \\
    & \textbf{Style} & TI & TC & EI & TI & TC & EI & TI & TC & EI & TI & TC & EI & TI & TC & EI & TI & TC & EI & TI & TC & EI \\
    \midrule
    \multirow{7}{*}{Llama3-8B} & ChatIE
    & 33.7 & 7.3 & 13.8
    & 20.8 & 10.2 & \textbf{27.6}
    & 30.6 & 24.9 & 46.8
    & 8.6 & 3.2 & 11.3
    & 28.4 & 15.5 & \textbf{43.3}
    & 10.8 & 3.6 & 20.4
    & 22.2 & 10.8 & 27.2
    \\
    & GEE
    & 19.1 & 1.9 & 6.8 
    & 11.7 & 5.9 & 14.0
    & 30.0 & 21.3 & 27.4
    & 25.4 & 15.8 & 26.7
    & \textbf{35.9} & 27.7 & 38.7 
    & 11.5 & 9.2 & 45.8
    & 22.3 & 13.6 & 26.6
    \\
    & DEE
    & 33.7 & 6.0 & 9.2 
    & 21.1 & 10.6 & 17.8 
    & 26.9 & 19.8 & 36.1
    & 25.3 & 16.9 & \textbf{32.5}
    & 29.1 & 20.3 & 39.2
    & 8.7 & 7.6 & 48.3 
    & 24.1 & 13.5 & 30.5
    \\
    & BD 
    & \textbf{54.5} & 10.7 & 12.3
    & 22.3 & 9.9 & 15.0 
    & 34.2 & 19.5 & 31.4 
    & \textbf{28.1} & 11.2 & 30.2 
    & 35.3 & 24.7 & 37.2 
    & 16.8 & 7.4 & 44.5 
    & 31.9 & 13.9 & 28.4
    \\
    & MD
    & 45.9 & 2.8 & 4.0
    & 25.2 & 9.5 & 15.2
    & 35.6 & 22.4 & 30.1
    & 22.8 & 15.3 & 25.4
    & 34.9 & \textbf{27.8} & 42.4
    & 10.3 & 8.8 & 47.9
    & 29.1 & 14.4 & 27.5 
    \\
    & MS
    & 46.2 & 10.3 & 11.2
    & 20.2 & 10.2 & 17.0 
    & 26.7 & 17.6 & 23.1 
    & 27.6 & \textbf{19.7} & 30.5 
    & 34.1 & 27.3 & 40.6
    & 11.9 & 10.3 & 48.3 
    & 27.8 & 15.9 & 28.4 
    \\
    & \modelName
    & 53.5 & \textbf{14.4} & \textbf{17.4} 
    & \textbf{26.1} & \textbf{15.7} & 25.0 
    & \textbf{40.3} & \textbf{36.3} & \textbf{47.9} 
    & 25.8 & 15.4 & 30.0 
    & 35.5 & 23.6 & 42.4 
    & \textbf{18.5} & \textbf{16.8} & \textbf{58.8}
    & \textbf{33.3} & \textbf{20.4} & \textbf{36.9}
    \\
    \midrule
    \multirow{7}{*}{Llama3-70B} & ChatIE
    & 47.9 & 19.8 & 24.8
    & 33.3 & 20.8 & \textbf{40.6}
    & 45.5 & 37.9 & 47.0
    & 14.6 & 6.4 & 17.3
    & 41.8 & 31.0 & 50.9
    & 12.9 & 10.2 & 48.9
    & 32.7 & 21.0 & 38.3
    \\
    & GEE
    & 28.3 & 15.7 & 17.5
    & 26.2 & 16.3 & 31.1
    & 47.0 & 42.3 & 52.2
    & 32.5 & 24.2 & 38.5
    & 43.7 & 34.7 & 46.0
    & 11.1 & 10.7 & 43.2
    & 31.5 & 24.0 & 38.1
    \\
    & DEE
    & 60.8 & 14.8 & 16.4
    & 34.0 & 21.3 & 33.6
    & 47.4 & 38.3 & 45.4
    & 39.2 & 30.5 & 46.0
    & 41.7 & 32.2 & 44.7
    & 16.6 & 16.4 & 63.1
    & 40.0 & 25.6 & 41.5
    \\
    & BD
    & 63.0 & 13.9 & 15.2 
    & 34.0 & 14.5 & 22.6 
    & 49.1 & 36.6 & 41.7 
    & \textbf{39.4} & 26.5 & 45.4 
    & \textbf{49.2} & 33.6 & 45.7 
    & 16.5 & 11.7 & 48.8 
    & 41.9 & 22.8 & 36.6
    \\
    & MD
    & \textbf{63.5} & 14.2 & 14.7
    & 34.0 & 20.9 & 32.6
    & 51.2 & 40.2 & 46.8
    & 36.8 & 28.9 & 43.0
    & 45.4 & 36.8 & 49.0
    & 13.9 & 13.7 & 64.4 
    & 40.8 & 25.8 & 41.8
    \\
    & MS
    & 33.9 & 21.6 & 22.3 
    & 35.3 & 24.9 & 39.9
    & 49.9 & 42.8 & 46.9
    & 37.4 & \textbf{31.0} & 45.0
    & 43.8 & 35.5 & 49.6 
    & 14.0 & 14.0 & 59.5
    & 35.7 & 28.3 & 43.9 
    \\
    & \modelName
    & 62.5 & \textbf{27.8} & \textbf{30.6}
    & \textbf{40.4} & \textbf{25.1} & 36.1 
    & \textbf{57.2} & \textbf{49.5} & \textbf{55.1} 
    & 38.6 & \textbf{31.0} & \textbf{48.5}
    & 45.0 & \textbf{36.5} & \textbf{51.8}
    & \textbf{17.3} & \textbf{16.6} & \textbf{66.6}
    & \textbf{43.5} & \textbf{32.8} & \textbf{48.1}
    \\
    \bottomrule
\end{tabular}}
\caption{Main results comparing the zero-shot ED performance of our proposed \modelName{} with all other baselines for the Llama3-8B-Instruct and Llama3-70B-Instruct LLMs. TI: Trigger Identification, TC: Trigger Classification, EI: Event Identification. \textbf{bold} = best performance. (XX) = number of distinct event types.}
\label{tab:llama-zd-results}
\end{table*}

\paragraph{FSM-guided decoding for constraint enforcement:}
To reduce the burden of constraint-following on the LLM and ensure strict adherence to the task format, we incorporate a constrained decoding mechanism guided by a finite-state machine (FSM).
Inspired by recent work \cite{outlines, ctrl-g}, the FSM explicitly encodes structural and task-specific constraints (Eq.~\ref{eq:json-constraint}–\ref{eq:event-constraint}) within the decoding process.
We construct and demonstrate an FSM to encode constraints for our ED task in Figure~\ref{fig:grounder-fsm}.

The FSM states represent decision points (e.g., whether the sentence contains an event, which event type $e\in\mathcal{E}$ to choose, which trigger $w\in X$ to assign, etc.), and the transitions denote valid LLM generations at each point (e.g., list of event types in $\mathcal{E}$, trigger words in the sentence).
As shown in Figure~\ref{fig:grounder-fsm}, generation proceeds step by step: starting with the event/no-event decision (state A), followed by selecting an event type (state B), then its trigger (state C), and finally deciding whether to generate another event mention or terminate (state D).
To ensure that the generations are natural, the FSM states are partitioned in alignment with the LLM tokenizer, i.e., the states are chosen such that the sequence of transition tokens is the most probable tokenization of the output text $Y_g$.

We implement this FSM using the Outlines library \cite{outlines} integrated into a vLLM inference framework \cite{vllm}.
The module takes as input the ontology, input sentence, LLM, and output JSON schema (potentially expressed as a grammar).
Each FSM state transition is encoded as an Outlines \texttt{choices} list, thereby restricting the LLM’s output vocabulary to only valid strings for that transition.
For example, the set of possible event types or candidate trigger words is directly provided as the restricted vocabulary, and transitions with a single option are handled deterministically.
The selected string then determines the next FSM state.

This design enforces structural validity during decoding: at each step, tokens not corresponding to valid FSM transitions are zeroed out, ensuring the LLM can only generate ontology-compliant outputs. Our implementation currently supports generation of JSON tuples of the form (event type, trigger), making it directly applicable to any ED dataset. More generally, because the transition and state mappings can be automatically constructed from the grammar of task constraints, the approach is customizable to other output formats and structured prediction tasks.

\subsection{Judge}
\label{sec:judge}

The final component of our pipeline, Judge \textit{aka High precision verifier}, serves to ensure each predicted event mention adheres to the original task instructions.
Specifically, for each candidate pair $(e_i, t_i)$, the Judge $f_j$ evaluates the hypothesis that the trigger $t_i$ expresses the event type $e_i$ in the context of the input sentence $X$ as
\begin{align}
    y^i_j = ``Yes/No" = f_j(e_i, t_i, X; \mathcal{E}) \nonumber
\end{align}
All predictions with $y^i_j = ``Yes"$ are accepted into the final output, while the others are rejected.
We provide an illustration of the prompt in Figure~\ref{fig:judge-prompt}.

This verification step plays a crucial role in ensuring the semantic validity and task alignment of predictions at a holistic level.
By filtering out irrelevant or uncertain outputs, the Judge substantially improves the precision of the overall system without requiring additional supervision or training.

\begin{table*}[t!]
\centering
\setlength{\tabcolsep}{3.5pt}
\resizebox{\textwidth}{!}{
\begin{tabular}{ll|ccc|ccc|ccc|ccc|ccc|ccc|ccc}
    \toprule
    \multirow{2}{*}{\textbf{LLM}} & \textbf{Prompt} & \multicolumn{3}{c|}{\textbf{MAVEN} (168)} & \multicolumn{3}{c|}{\textbf{FewEvent} (100)} & \multicolumn{3}{c|}{\textbf{ACE} (33)} & \multicolumn{3}{c|}{\textbf{GENIA} (9)} & \multicolumn{3}{c|}{\textbf{SPEED} (7)} & \multicolumn{3}{c|}{\textbf{CASIE} (5)} & \multicolumn{3}{c}{\textbf{Average}} \\
    & \textbf{Style} & TI & TC & EI & TI & TC & EI & TI & TC & EI & TI & TC & EI & TI & TC & EI & TI & TC & EI & TI & TC & EI \\
    \midrule
    \multirow{3}{*}{Qwen2.5-14B} & MD
    & 53.0 & 17.6 & 20.9
    & 28.8 & \textbf{21.1} & \textbf{34.2} 
    & 28.3 & 24.5 & 42.1 
    & 24.8 & 18.8 & 26.7 
    & 37.7 & 33.0 & \textbf{51.2} 
    & 15.8 & 15.8 & \textbf{61.5} 
    & 31.4 & 21.8 & 39.5 
    \\
    & MS
    & 46.5 & 20.8 & 24.6
    & 24.8 & 18.9 & 32.1
    & 33.6 & 26.3 & 32.5
    & 25.4 & 19.2 & 27.7 
    & 38.9 & 34.3 & 46.1 
    & 16.3 & 16.1 & 54.5 
    & 30.9 & 22.6 & 36.2 
    \\
    & \modelName
    & \textbf{53.1} & \textbf{23.3} & \textbf{27.6}
    & \textbf{29.7} & 19.3 & 30.4
    & \textbf{38.4} & \textbf{37.7} & \textbf{48.8}
    & \textbf{29.9} & \textbf{22.6} & \textbf{38.6} 
    & \textbf{42.9} & \textbf{35.3} & 46.5 
    & \textbf{19.7} & \textbf{19.5} & 58.8
    & \textbf{35.8} & \textbf{26.1} & \textbf{41.8}
    \\
    \midrule
    \multirow{3}{*}{Qwen2.5-72B} & MD
    & 49.4 & 21.6 & 24.1 
    & 17.0 & 12.3 & 21.0 
    & 28.8 & 25.8 & 30.3 
    & 30.5 & 27.0 & 36.3
    & \textbf{41.4} & \textbf{37.4} & \textbf{45.4}
    & 11.0 & 10.4 & 57.9
    & 29.7 & 22.4 & 35.8 
    \\
    & MS
    & 39.9 & 23.6 & 25.4
    & 25.0 & 21.0 & \textbf{34.2}
    & 42.5 & 40.4 & 42.5
    & 26.7 & 23.6 & 34.1
    & 40.6 & 35.5 & 45.2
    & 10.5 & 10.5 & 49.1
    & 30.9 & 25.8 & 38.4
    \\
    & \modelName
    & \textbf{54.1} & \textbf{27.5} & \textbf{30.2}
    & \textbf{30.8} & \textbf{22.3} & 32.9
    & \textbf{46.8} & \textbf{44.8} & \textbf{47.8} 
    & \textbf{33.6} & \textbf{29.8} & \textbf{43.9}
    & 40.6 & 34.7 & 41.4
    & \textbf{15.9} & \textbf{15.8} & \textbf{59.3}
    & \textbf{37.0} & \textbf{29.2} & \textbf{42.6}
    \\
    \midrule
    \multirow{3}{*}{GPT3.5-turbo} & MD
    & \textbf{50.9} & 17.4 & 20.4
    & 23.2 & 14.6 & 27.0 
    & 40.9 & 36.2 & 42.5
    & \textbf{27.0} & \textbf{19.9} & 31.4
    & \textbf{36.5} & \textbf{30.6} & 41.8
    & 10.0 & 9.9 & 51.1
    & \textbf{31.4} & 21.4 & 35.7
    \\
    & MS
    & 48.2 & 15.5 & 17.2 
    & 23.7 & \textbf{15.9} & 29.8
    & 40.7 & 37.4 & 42.3
    & 23.2 & 19.0 & 26.3
    & 33.0 & 23.7 & 35.5 
    & 7.7 & 7.1 & 44.4
    & 29.4 & 19.8 & 32.6
    \\
    & \modelName
    & 48.1 & \textbf{21.6} & \textbf{26.1}
    & \textbf{25.3} & 15.6 & \textbf{31.1}
    & \textbf{41.7} & \textbf{41.7} & \textbf{48.9}
    & 26.2 & 19.5 & \textbf{36.3}
    & 32.4 & 27.2 & \textbf{49.0}
    & \textbf{11.4} & \textbf{10.6} & \textbf{55.7}
    & 30.9 & \textbf{22.7} & \textbf{41.2}
    \\
    \midrule
    \multirow{3}{*}{GPT4o} & MD
    & \textbf{61.8} & 28.9 & 31.9
    & 30.6 & 23.9 & 35.4
    & 52.3 & 52.3 & 52.3
    & \textbf{41.0} & \textbf{36.5} & 49.5 
    & 44.1 & 40.2 & 48.0 
    & 10.1 & 10.1 & 55.7
    & 40.0 & 32.0 & 45.5
    \\
    & MS
    & 49.4 & 30.8 & 33.3 
    & 25.6 & 20.6 & 32.2 
    & 36.2 & 36.2 & 38.3 
    & 36.6 & 33.2 & 45.0
    & \textbf{45.7} & \textbf{40.4} & \textbf{50.1}
    & 13.4 & 13.4 & 46.9
    & 34.5 & 29.1 & 41.0
    \\
    & \modelName
    & 58.5 & \textbf{32.2} & \textbf{35.6} 
    & \textbf{36.1} & \textbf{28.4} & \textbf{38.5}
    & \textbf{54.9} & \textbf{54.9} & \textbf{56.6} 
    & 40.7 & 35.4 & \textbf{51.2} 
    & 43.3 & 37.3 & 46.1 
    & \textbf{16.7} & \textbf{16.7} & \textbf{58.8} 
    & \textbf{41.7} & \textbf{34.2} & \textbf{47.8}
    \\
    \bottomrule
\end{tabular}}
\caption{Generalization results for zero-shot ED performance comparing \modelName{} with the best baselines for four other LLMs of Qwen2.5-14B-Instruct, Qwen2.5-72B-Instruct, GPT3.5-turbo, and GPT4o. \textbf{bold} = best performance. (XX) = number of distinct event types.}
\label{tab:other-llms-zd-results}
\end{table*}

\section{Experimental Setup}
\label{sec:expt}

In this section, we describe our experimental setup comprising the datasets, baselines, evaluation metrics, and implementation details.
Additional setup and implementation details are provided in \S~\ref{sec:appendix-expt-details}.

\paragraph{Datasets:}
We benchmark our model across six ED datasets spanning five diverse domains, listed as:
(1) MAVEN \cite{wang-etal-2020-maven} and
(2) FewEvent \cite{fewevent} from the general domain,
(3) ACE \cite{doddington-etal-2004-automatic} from the news domain,
(4) GENIA \cite{kim-etal-2011-overview-genia}, from the biomedical domain,
(5) SPEED \cite{parekh-etal-2024-event}, from the epidemiological/social media domain,
(6) CASIE \cite{casie}, from the cybersecurity domain.

\begin{table}[t]
    \centering
    \small
    \setlength{\tabcolsep}{3.5pt}
    \begin{tabular}{llccc}
        \toprule
        \textbf{Dataset} & \textbf{Domain} & \textbf{\# Doc} & \textbf{\# Event} & \textbf{Avg. Doc} \\
        & & & \textbf{Mentions} & \textbf{Length} \\
        \midrule
        MAVEN & General & 250 & 623 & 24.5 \\
        FewEvent & General & 250 & 250 & 30.5 \\
        ACE & News & 250 & 71 & 13.2 \\
        GENIA & Biomedical & 250 & 2472 & 251.3 \\
        SPEED & Epidemiology & 250 & 258 & 32.4 \\
        CASIE & Cybersecurity & 50 & 291 & 283.1 \\
        \bottomrule
    \end{tabular}
    \caption{Data Statistics of the various ED datasets used in our experimental setup.}
    \label{tab:data-statistics}
\end{table}

We provide statistics about the test splits of the different datasets in Table~\ref{tab:data-statistics}.
To avoid any distributional biases, following TextEE \cite{huang-etal-2024-textee}, we uniformly sample 250 datapoints from the combined train-dev-test splits of each dataset for evaluation.
Since CASIE is a smaller dataset, we only use 50 test samples for this dataset.
The table highlights the domain diversity of the datasets covering common domains like news and general, while also focusing on technical domains like biomedical and epidemiology.
The datasets also show variation in the density, with ACE, FewEvent, and SPEED being sparse with upto 1 event mention/sentence.
On the other hand, MAVEN, CASIE, and GENIA are denser with 2.5-10 event mentions/passage.
Finally, we also show the variation in token length, with ACE being the lowest with an average of 13 tokens, while GENIA and CASIE are longer with 250-280 average tokens per input document.

\paragraph{Baselines:}
We consider two major baselines, described below:
(1) Multi-event Direct (MD) \cite{chatgpt-ed} directly prompts the LLM to provide the final output in a single pass, and
(2) Multi-event Staged (MS) \cite{fig} decomposes the task into two stages, where the first stage identifies the event and the second stage extracts the corresponding triggers.
We also compare with other works like:
(3) Binary-event Direct (BD) \cite{lyu-etal-2021-zero, li-etal-2023-glen} prompts the LLM to do binary classification for each event,
(4) Decompose-Enrich-Extract (DEE) \cite{dee} utilizes instruction enrichment with schema information for ED,
(5) GuidelineEE (GEE) \cite{guideline-ee}, similar to Code4Struct \cite{wang-etal-2023-code4struct}, converts ED into a code-generation problem using Python classes and instantiations, and
(6) ChatIE \cite{chatie} decomposes ED via multi-turn conversations.
We ensure consistent, structured outputs for each baseline to maintain fair comparisons (analysis in \S~\ref{sec:appendix-structured-output-analysis}).
Furthermore, we add the Judge component to each baseline, if not already present, to ensure robust benchmarking of \modelName.

\paragraph{Base LLMs:}
We use the following LLMs for our base experiments: Llama3-8B-Instruct and Llama3-70b-Instruct from the Llama3 family \cite{llama3} and Qwen2.5-14B-Instruct; Qwen2.5-72B-Instruct from the Qwen2.5 \cite{qwen} LLM family; and GPT3.5-turbo and GPT-4o \cite{gpt, gpt4} from OpenAI.

\paragraph{Evaluation Metrics:}
Following \citet{ahn-2006-stages, fig} we report the F1 scores for the following three metrics:
(1) Trigger Identification (TI) - correct identification of triggers, and
(2) Event Identification (EI) - correct classification of event types, and
(3) Trigger Classification (TC) - correct identification of the trigger-event pair (event mention).
To maintain consistency with traditional span-based evaluations, we used string matching to map the generated outputs to input spans.

\paragraph{Implementation Details:}
\label{sec:implementation-details}
We use TextEE \cite{huang-etal-2024-textee} for our benchmarking, datasets, and evaluation setup.
To restrict LLM's generation choices for the FSM-guided constrained decoding, we utilize Outlines \cite{outlines} over vLLM inference \cite{vllm}.
We use Curator \cite{curator} for querying the GPT family LLMs.
We deploy a temperature of 0.4 and top-p of 0.9 for decoding.
We report the averaged results over three runs for robust benchmarking.

\section{Results and Analysis}
\label{sec:results}

In this section, we provide our main results and findings, and later provide supporting evidence through our analyses.
We also provide additional experimental results and error anlaysis in the Appendix (\S~\ref{sec:appendix-results}).

\subsection{Main Results}
\label{sec:main-results}

We present the main zero-shot results for all baselines on the six datasets for Llama3 LLMs in Table~\ref{tab:llama-zd-results}.
As seen from the average results (last three columns), \modelName{} performs the best, surpassing the best baseline of multi-event staged (MS) by a significant margin of 5.5-8\% TI, 4-8.5\% EI, and 4-5\% TC.
The performance disparity across different task decomposition methods of ChatIE, MS, and \modelName{} highlights how our divergent-convergent decomposition of Dreamer-Grounder provides a stronger inductive bias.
Other baselines perform relatively better on datasets like GENIA/SPEED, as these are simpler datasets with fewer event types; thus, requiring lesser cognitive reasoning.
However, on the high-event datasets like MAVEN/FewEvent/ACE which require more complex reasoning, \modelName{} with its divergent-convergent reasoning shows more significant improvement over the baselines.

\begin{table}[t]
    \centering
    \small
    \begin{tabular}{l|ccc}
        \toprule
        \textbf{Model Setting} & \multicolumn{3}{c}{\textbf{Average F1}} \\
        & TI & TC & EI\\
         \midrule
        \multicolumn{4}{c}{\textbf{Test on GENIA, SPEED, CASIE}} \\
        \midrule
        GOLLIE-7B & 6.0 & 5.3 & 15.3 \\
        GOLLIE-34B & 15.6 & 11.7 & 29.4 \\
        Llama3-8B \modelName & \underline{26.6} & \underline{18.6} & \underline{43.7} \\
        Llama3-70B \modelName & \underline{33.6} & \underline{28.0} & \underline{55.6} \\
        \midrule
        \multicolumn{4}{c}{\textbf{Test on all but ACE dataset}} \\
        \midrule
        ACE-trained DEGREE & 20.9 & 11.0 & 21.3 \\
        Llama3-8B \modelName & \underline{31.9} & \underline{17.2} & \underline{34.7} \\
        Llama3-70B \modelName & \underline{40.8} & \underline{27.4} & \underline{46.7} \\
        \midrule
        \multicolumn{4}{c}{\textbf{Test on all but MAVEN dataset}} \\
        \midrule
        MAVEN-trained DEGREE & 31.8 & 25.0 & 38.6 \\
        Llama3-8B \modelName & 29.2 & 21.6 & \underline{40.8} \\
        Llama3-70B \modelName & \underline{39.7} & \underline{31.7} & \underline{51.6} \\
        \bottomrule
    \end{tabular}
    \caption{Comparison of pure zero-shot \modelName{} with fine-tuned transfer-learning baselines. \underline{Underline} indicates scenarios of \modelName{} improvements.}
    \label{tab:transfer-learning-results}
\end{table}

\paragraph{Generalization across LLMs:}
To demonstrate the generalizability of \modelName, we benchmark it with the top-performing baselines on four additional LLMs from the Qwen and GPT families and show our results in Table~\ref{tab:other-llms-zd-results}.
We note how \modelName{} performs the best across all LLMs with an overall average improvement of 5.5\% TI, 6.5\% EI, 4\% TC over the multievent-staged baseline and 3.3\% TI, 5.4\%, 4.6\% TC over the multievent-direct baseline.
Across different LLMs, we note the strongest performance on GPT4o, followed by Llama3-70B-Instruct and Qwen2.5-72B, indicating how more parameters help better reasoning with \modelName.

\begin{table}[t]
    \centering
    \small
    \setlength{\tabcolsep}{4.5pt}
    \begin{tabular}{ll|ccc}
        \toprule
        \textbf{Base LLM} & \textbf{Prompt} & \multicolumn{3}{c}{\textbf{Average F1}} \\
        & \textbf{Style} & TI & TC & EI\\
        \midrule
        \multicolumn{5}{c}{\textbf{Chain-of-thought Baselines}} \\
        \midrule
        Llama3-8B & MD + CoT & 25.0 & 13.5 & 27.1 \\
        Llama3-8B & MS + CoT & 28.4 & 17.6 & 31.9 \\
        Llama3-70B & MD + CoT & 41.0 & 30.9 & 48.0 \\
        Llama3-70B & MS + CoT & 40.5 & 31.6 & 47.1 \\
        Qwen2.5-72B & MD + CoT & 34.9 & 27.1 & 43.6 \\
        Qwen2.5-72B & MS + CoT & 36.2 & 28.8 & 40.8 \\
        \midrule
        \multicolumn{5}{c}{\textbf{Thinking-based model Baselines}} \\
        \midrule
        DS-Qwen-32B & MD & 39.2 & 30.0 & 46.3 \\
        DS-Qwen-32B & MS & 39.5 & 30.4 & 45.2 \\
        DS-Llama3-70B & MD & 29.0 & 23.3 & 36.1 \\
        DS-Llama3-70B & MS & 33.3 & 27.0 & 37.8 \\
        O1-mini & MD & 40.2 & 32.5 & 44.7 \\
        \midrule
        \multicolumn{5}{c}{\textbf{\modelName{} base model results}} \\
        \midrule
        Llama3-8B & \modelName & \underline{33.3} & \underline{20.4} & \underline{36.9} \\
        Llama3-70B & \modelName & \underline{43.5} & \underline{32.8} & \underline{48.1} \\
        Qwen2.5-72B & \modelName & \underline{37.0} & \underline{29.2} & 42.6 \\
        GPT4o & \modelName & \underline{41.7} & \underline{34.2} & \underline{47.8} \\
        \midrule
        \multicolumn{5}{c}{\textbf{\modelName{} improvements with reasoning}} \\
        \midrule
        Llama3-8B & \modelName + CoT & 33.1 & 21.1 & 36.2 \\
        Llama3-70B & \modelName + CoT & 43.0 & 33.1 & \textbf{49.8} \\
        Qwen2.5-72B & \modelName + CoT & 37.0 & 29.1 & 43.5 \\
        DS-Qwen-32B & \modelName & \textbf{43.1} & \textbf{33.3} & 49.5 \\
        DS-Llama3-70B & \modelName & 41.4 & 33.0 & 48.3 \\
        \bottomrule
    \end{tabular}
    \caption{Comparison of \modelName{} with reasoning-based baselines like Chain-of-thought (CoT) and thinking-based models. \underline{Underline} indicates \modelName{} improvements over reasoning baselines.}
    \label{tab:reasoning-results}
\end{table}

\begin{table}[t]
    \centering
    \small
    \setlength{\tabcolsep}{3.8pt}
    \begin{tabular}{l|ccc|ccc}
        \toprule
        \textbf{Component} & \multicolumn{3}{c|}{\textbf{TI}} & \multicolumn{3}{c}{\textbf{TC}} \\
        & P & R & F & P & R & F \\
        \midrule
        \multicolumn{7}{c}{\textbf{Llama3-8B-Instruct}} \\
        \midrule
        Dreamer & 8.5 & \textbf{64.3} & 15.0 & 0.0 & 0.0 & 0.0 \\
        \quad + Grounder & 20.4 & 47.9 & 28.6 & 15.5 & 37.1 & 21.9 \\
        \quad + FSM Decoding & 22.3 & 56.8 & 32.1 & 16.2 & \textbf{42.3} & 23.4 \\
        \quad + Judge & 41.8 & 39.0 & \textbf{40.3} & \textbf{37.5} & 35.2 & \textbf{36.3} \\
        \midrule
        MD Baseline & \textbf{48.4} & 28.2 & 35.6 & 30.2 & 17.8 & 22.4 \\
        MS Baseline & 22.0 & 33.8 & 26.7 & 14.4 & 22.5 & 17.6 \\
        \midrule
        \multicolumn{7}{c}{\textbf{Llama3-70B-Instruct}} \\
        \midrule
        Dreamer & 15.5 & \textbf{77.5} & 25.8 & 0.0 & 0.0 & 0.0 \\
        \quad + Grounder & 28.6 & 65.7 & 40.4 & 22.5 & 53.4 & 31.8 \\
        \quad + FSM Decoding & 32.3 & 66.7 & 43.5 & 26.2 & 54.0 & 35.3 \\
        \quad + Judge & 52.8 & 62.5 & \textbf{57.2} & 45.7 & \textbf{54.0} & \textbf{49.5} \\
        \midrule
        MD Baseline & 57.2 & 46.5 & 51.2 & 44.0 & 37.1 & 40.2 \\
        MS Baseline & \textbf{66.4} & 39.9 & 49.9 & \textbf{57.0} & 34.3 & 42.8 \\
        \bottomrule
    \end{tabular}
    \caption{Ablation Study on the ACE dataset highlighting the significance and contribution of each component of \modelName. P: Precision, R: Recall, F: F1 score.}
    \label{tab:ablation-study}
\end{table}

\begin{table*}[t]
    \centering
    \small
    \begin{tabular}{p{3cm}|p{2.5cm}|p{2.8cm}|p{3cm}|p{2.5cm}}
        \toprule
        \textbf{Sentence} & \textbf{Best Baseline} & \textbf{Dreamer } & \textbf{Grounder} & \textbf{Judge} \\
        & \textbf{Prediction} & \textbf{Prediction} & \textbf{Prediction} & \textbf{Prediction} \\
        \midrule
        cass apd ra gave birth to her first daughter. & [(\greentext{"Life:Be-Born"}, \redtext{"gave"})] & [("Birth", "gave"), ("Birth", "birth")] & [("Life:Be-Born", "birth")] & [(\greentext{"Life:Be-Born", "birth"})] \\
        \midrule
        After passing the island, the hurricane turned to the northeast, and became extratropical on September 8, before dissipating two days later. & [(\redtext{"Change", "turned"}), (\redtext{"Change", "became"}), (\redtext{"Dissipating"}, \greentext{"dissipating"})] & [("Movement", "turned"), ("Transition", "became"), ("Dissipation", "dissipating")] & [("Change\_event\_time", "turned"), ("Becoming\_a\_member", "became"), ("Dispersal", "dissipating")] & [(\greentext{"Dispersal", "dissipating"})] \\
        \midrule
        Covid-19 has led to social distancing, but we can still be together through the quarantine with online gaming! & \redtext{[]} & [("Social\_Distancing", "distancing"), ("Quarantine", "quarantine"), ("Gaming", "gaming")] & [("prevent", "distancing"), ("control", "quarantine")] & [(\greentext{"prevent", "distancing")}, (\greentext{"control", "quarantine"})] \\
        \bottomrule
    \end{tabular}
    \caption{Qualitative examples comparing \modelName's predictions (per component) with the best baseline. We highlight the correct predictions in \greentext{green} and incorrect ones in \redtext{red}.}
    \label{tab:qual-study}
\end{table*}

\subsection{Comparison with Fine-tuned Transfer-learning Methods}
\label{sec:transfer-learning}

Various works utilize transfer-learning and universal Information Extraction (IE) training for zero-shot cross-dataset ED \cite{cai-etal-2024-improving-event, li-etal-2024-knowcoder}.
These works train on selected IE datasets and show performance on unseen IE datasets.
We provide a comparison of \modelName{} with two such transfer-learning approaches:
(1) DEGREE \cite{hsu-etal-2022-degree}, a generative framework utilizing text-based event templates to generalize,
(2) GOLLIE \cite{gollie}, a universal IE framework, fine-tuning LLMs on various IE instruction datasets.
For DEGREE, we consider two versions where the source data is ACE and MAVEN, respectively.
For GOLLIE, we consider the fine-tuned GOLLIE-7B and GOLLIE-34B models.
We provide the averaged results across target datasets (not included in the source data) in Table~\ref{tab:transfer-learning-results}, with detailed results in \S~\ref{sec:appendix-transfer-learning}.
Through these results, we demonstrate how, despite no fine-tuning, \modelName{} consistently outperforms the fine-tuned transfer-learning baselines across all settings.
On average, \modelName{} improves by 3-10\% F1 using Llama3-8B-Instruct and 10-22\% F1 using Llama3-70B-Instruct and GPT4o.

\subsection{Comparison with Reasoning baselines}
\label{sec:reasoning-analysis}

Reasoning by verbalizing thoughts using additional tokens has commonly helped improve performance across a wide range of tasks \cite{zs-cot, o1-assessment}.
We evaluate the utility of reasoning, specifically Chain-of-thought (CoT) \cite{cot}, along with thinking-based models like Deepseek-R1-Distilled-Qwen-32B (DS-Qwen-32B), Deepseek-R1-Distilled-Llama3-70B (DS-Llama3-70B) \cite{deepseek-r1} and O1-mini \cite{o1} on our task of zero-shot ED in Table~\ref{tab:reasoning-results} (complete results in \S~\ref{sec:appendix-reasoning-results}).
We demonstrate how the baselines improve with additional reasoning; however, \modelName{} with the base models (Llama3-70B) consistently outperforms all these reasoning baselines (even O1-mini) while using 15-55x fewer tokens on average (\S~\ref{sec:appendix-reasoning-results}).
We also show how our method is complementary to reasoning by demonstrating further improvements up to 1-2\% F1 using reasoning with \modelName.

\subsection{Ablation Study}
\label{sec:ablation-study}

To demonstrate the role of each component of our pipeline, we ablate and show the model performance as we add each component in \modelName{} for the ACE dataset for Llama3-8B and Llama3-70B LLMs in Table~\ref{tab:ablation-study}.
For reference, we also show the precision/recall splits of the baselines.
Dreamer achieves a high recall for TI (albeit a low precision) - demonstrating the utility of divergent unconstrained reasoning.
Grounder helps align the predictions, causing a slight drop in recall while improving the precision.
FSM Decoding helps largely improve the recall for Llama3-8B-Instruct by improving the mapping, and precision for Llama3-70B-Instruct by fixing any constraint violations.
Finally, Judge largely boosts the precision of the model.
Analysis of the baselines reveals that they are conservative, making a low number of high-precision predictions.
In comparison, \modelName{} makes many more predictions, largely improving recall while maintaining reasonably high precision.

\paragraph{Qualitative Study:}
We provide some qualitative examples for each component of \modelName{}, while comparing the best baseline across the datasets in Table~\ref{tab:qual-study} (more examples in \S~\ref{sec:appendix-qual-study}).
We see how the best baseline often reasons incorrectly, leading to precision loss, or remains conservative, predicting nothing, leading to recall errors.
The split across the three components shows how Dreamer generates many plausible event mentions, Grounder aligns and removes some, while Judge verifies and filters irrelevant ones.
These examples provide the internal workings of \modelName{}, highlighting the significance of divergent-convergent reasoning.

\section{Conclusion and Future Work}
\label{sec:conclusion}

In our work, we introduce \modelName, a novel divergent-convergent reasoning pipeline of Dreamer-Grounder-Judge, aimed at decoupling the LLM from task-specific constraints, and indirectly better exploiting LLMs' reasoning.
Through experimentation on six ED datasets from five domains across nine LLMs, we confirm how \modelName{} provides a stronger inductive bias, improving over other zero-shot baselines, fine-tuned transfer learning methods, and reasoning-focused approaches.
Future works can explore this paradigm on broader tasks and study to better elicit divergent-convergent reasoning.

\section*{Acknowledgments}

We express our gratitude to Po-Nien Kung and Haoyi Qiu for their valuable time, reviews of our work, and constructive feedback.
We thank the anonymous reviewers, area chairs, and the program committee for their reviews and feedback.
We are also grateful to Amazon for supporting this work through the Amazon Science Ph.D. Fellowship awarded to Tanmay Parekh.
This work was also partially supported by the National Science Foundation CAREER award \#2339766 and the Amazon AGI Research Award awarded to Nanyun Peng.

\section*{Limitations}
\label{sec:limitations}

In our work, we focus on improving zero-shot LLM inference for Event Detection.
This work is easily extendable to other low-resource settings as well as other Information Extraction (IE) tasks - but we leave these for future explorations.
To keep experimentation consistent with prior works, we utilized/sampled 250 datapoints from each dataset as our test set.
If working with a different data split, one might get different absolute model performance, but we believe the general trends should remain the same.
Finally, there are various lines of work on improving the use of retrieval to select good in-context examples, or teaching the LLM to learn the schema.
We believe these works are orthogonal and complementary to our work, and we do not compare/include them in our study.

\section*{Ethical Considerations}
\label{sec:ethical-considerations}

Our work focuses on using LLMs through the inductive bias of our method \modelName.
Since we do not train the LLM, there could be inherent biases in the LLM that can crop up when using our pipeline.
We do not study or provide methods to mitigate such biases, as it's not in the scope of our work.

We would like to acknowledge that we used AI assistants and chatbots for writing some parts of the paper, helping with coding up plots, and searching for related works.
For each application, a human expert verified to ensure we do not add any spurious/harmful content.

\bibliography{anthology,custom}

\clearpage

\appendix

\section{\modelName{} Prompts}
\label{sec:appendix-model-prompts}

We described our modeling paradigm of divergent-convergent reasoning through the Dreamer-Grounder-Judge paradigm in \S~\ref{sec:methodology}.
Here we provide some additional details and also share the prompts that we used for each component. 

\begin{figure}[t]
    \centering
    \includegraphics[width=\linewidth]{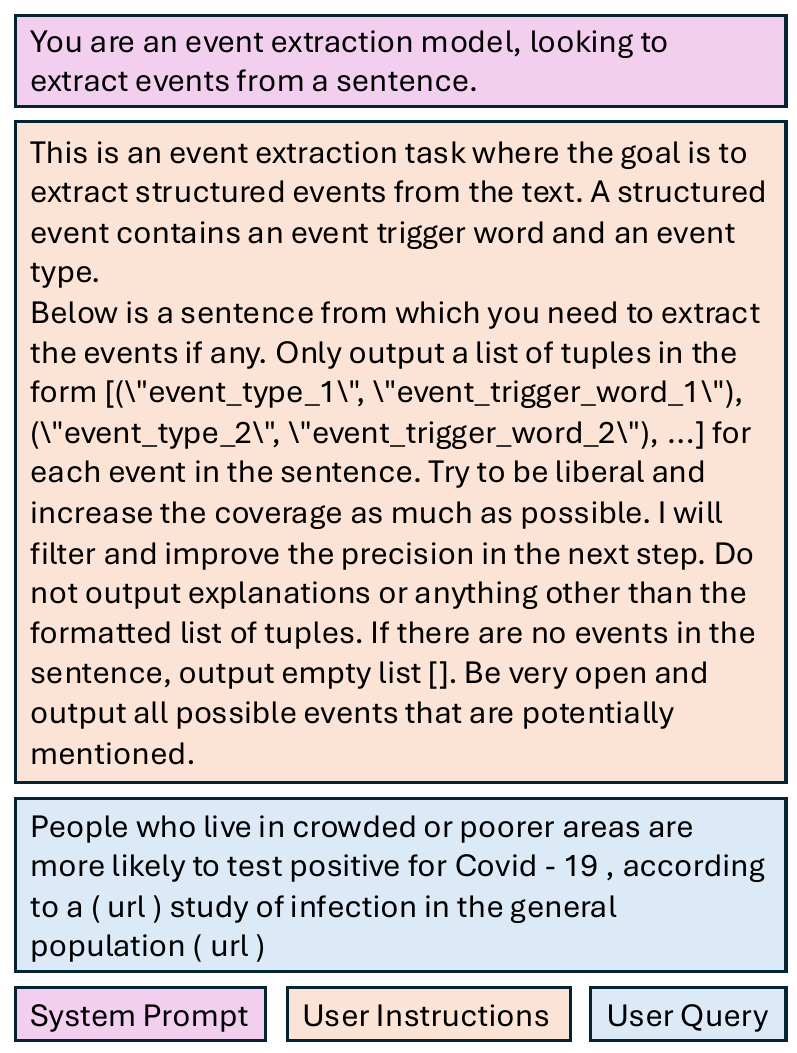}
    \caption{Illustration of the prompt utilized for Dreamer. To encourage divergent thinking, we remove event-based constraints from the model instructions. Furthermore, we add sentences that encourage the model to be liberal and open in its predictions.}
    \label{fig:dreamer-prompt}
\end{figure}

\paragraph{Dreamer:}
The Dreamer component induces divergent thinking, encouraging the model to think more widely.
We induce this behavior by removing the event-based constraints from the task instructions and adding additional inductive bias to provide this encouragement inthe  form of additional task instructions asking the model to be super liberal.
We provide an illustration of this prompt in Figure~\ref{fig:dreamer-prompt}.
Specifically, sentences like "Try to be liberal and increase the coverage as much as possible. I will filter and improve the precision in the next step." and "Be very open and output all possible events that are potentially mentioned." provide this stronger divergent reasoning inductive bias.

\begin{figure}[t]
    \centering
    \includegraphics[width=\linewidth]{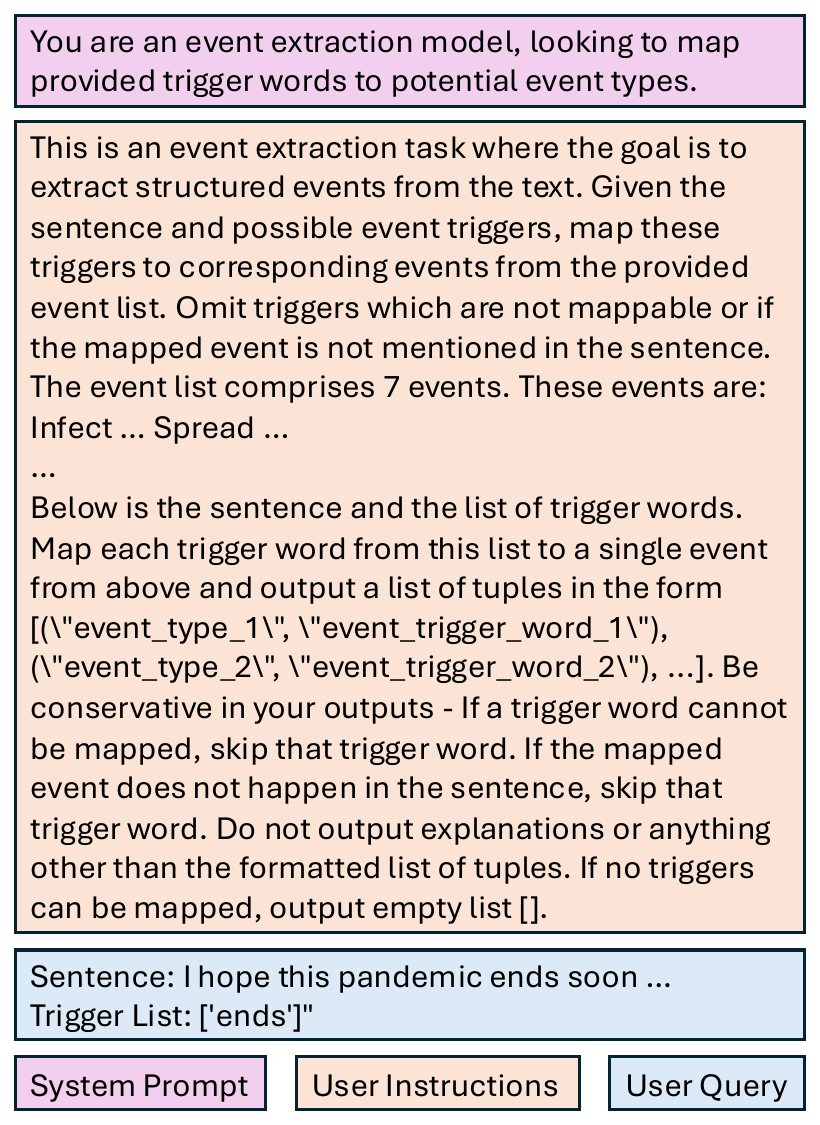}
    \caption{Illustration of the prompt utilized for Grounder. To encourage convergent thinking and alignment, we add event-based constraints in the model instructions. Furthermore, we add sentences that encourage the model to be more conservative in its predictions.}
    \label{fig:grounder-prompt}
\end{figure}

\paragraph{Grounder:}
The Grounder component aligns the open-ended predictions of the Dreamer with the closed event ontology using convergent reasoning.
To this end, we add the various task-specific constraints in the form of natural language instructions as well as use a finite-state machine (FSM) guided generation to aid with this convergent reasoning.
Here, we describe the prompt and the inductive biases in it, as illustrated in Figure~\ref{fig:grounder-prompt}.
Specifically, we first add all the verbalized constraints, including the ontology details in the form of event names and information.
To provide more inductive bias, we also add a sentence like "Be conservative in your outputs - If a trigger word cannot be mapped, skip the trigger word. If the mapped event does not happen in the sentence, skip the trigger word.".

\begin{figure}[t]
    \centering
    \includegraphics[width=\linewidth]{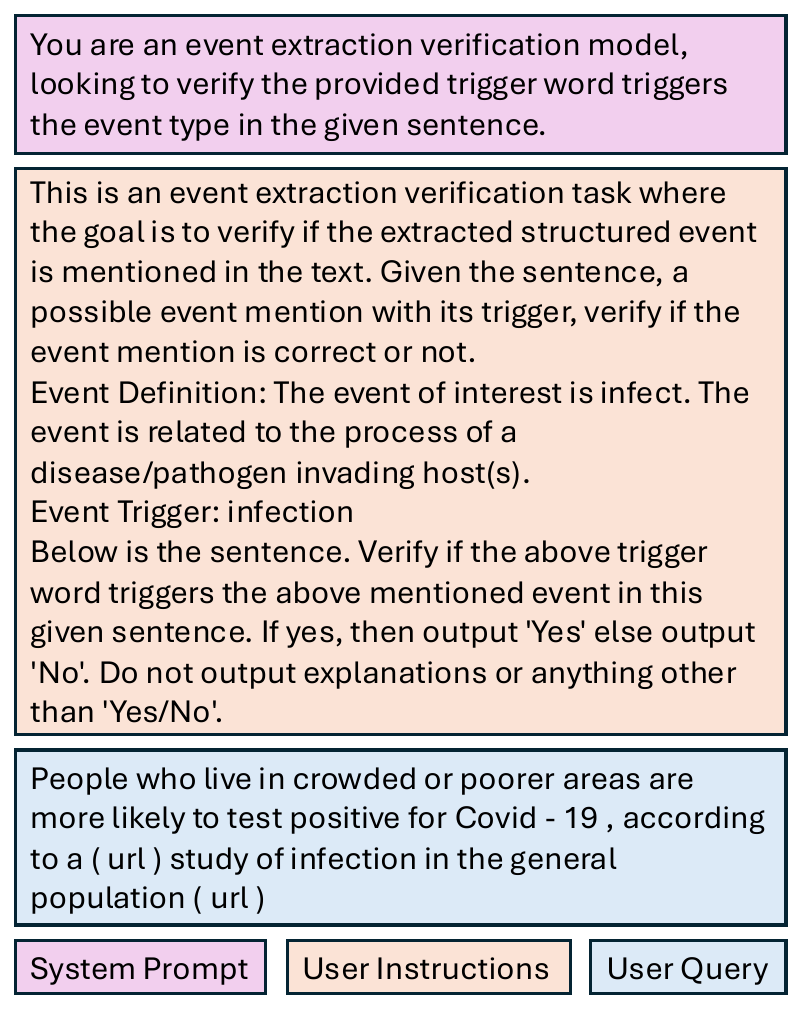}
    \caption{Illustration of the prompt utilized for Judge. To encourage convergent thinking and alignment, we add event-based constraints in the model instructions. Furthermore, we add sentences that encourage the model to be more conservative in its predictions.}
    \label{fig:judge-prompt}
\end{figure}

\paragraph{Judge:}
The Judge is tasked with the evaluation of the prediction to ensure that the trigger word triggers the specific event in the given sentence.
We run the Judge for each prediction separately.
To make this lightweight, we ensure that the output space is simple "Yes" or "No" without any explanation, which makes the parsing easier as well.
We provide an illustration of this prompt in Figure~\ref{fig:judge-prompt}.
This component is very generic and can be easily applied to other methods/LLMs as well.

\section{Additional Experimental Details}
\label{sec:appendix-expt-details}

In \S~\ref{sec:expt}, we provided brief details about our experimental and implementation details. Here, we provide additional implementation details for \modelName{} and the various baselines.
For open-source models, we ran them locally on NVIDIA RTX A6000/A100 machines with support for 8 GPUs.

\subsection{\modelName}

\paragraph{Trigger Atomization Adaptation for FSM-guided Decoding:}
\label{sec:appendix-trigger-atomization}
Different datasets have varied annotation instructions and definitions for the trigger spans.
Some datasets are strictly adhering to only single-word triggers (e.g., SPEED), while others are largely loose and support multi-word triggers (e.g., CASIE).
We provide a small study of measuring multi-word triggers in Table~\ref{tab:multi-trigger-analysis}, highlighting this disparity across datasets.
To account for these varied definitions, we infuse a customizable atomization unit in our FSM-guided decoding.
Specifically, state C from Figure~\ref{fig:grounder-fsm} is customizable wherein for stricter datasets (SPEED, ACE, FewEvent), we impose an additional constraint of single-word trigger, while for other datasets (CASIE, GENIA, MAVEN), we apply a looser constraint of substring match with the query sentence.

\begin{table}[h]
    \centering
    \small
    \begin{tabular}{lc}
        \toprule
        \textbf{Dataset} & \textbf{\% Multi-word Triggers} \\
        \midrule
        MAVEN & 8\% \\
        FewEvent & 3\% \\
        ACE & 2.8\% \\
        GENIA & 8.5\% \\
        SPEED & 0\% \\
        CASIE & 54.6\% \\
        \bottomrule
    \end{tabular}
    \caption{Measuring the percentage of multi-word triggers across the different ED datasets.}
    \label{tab:multi-trigger-analysis}
\end{table}

\begin{figure}
    \centering
    \includegraphics[width=\linewidth]{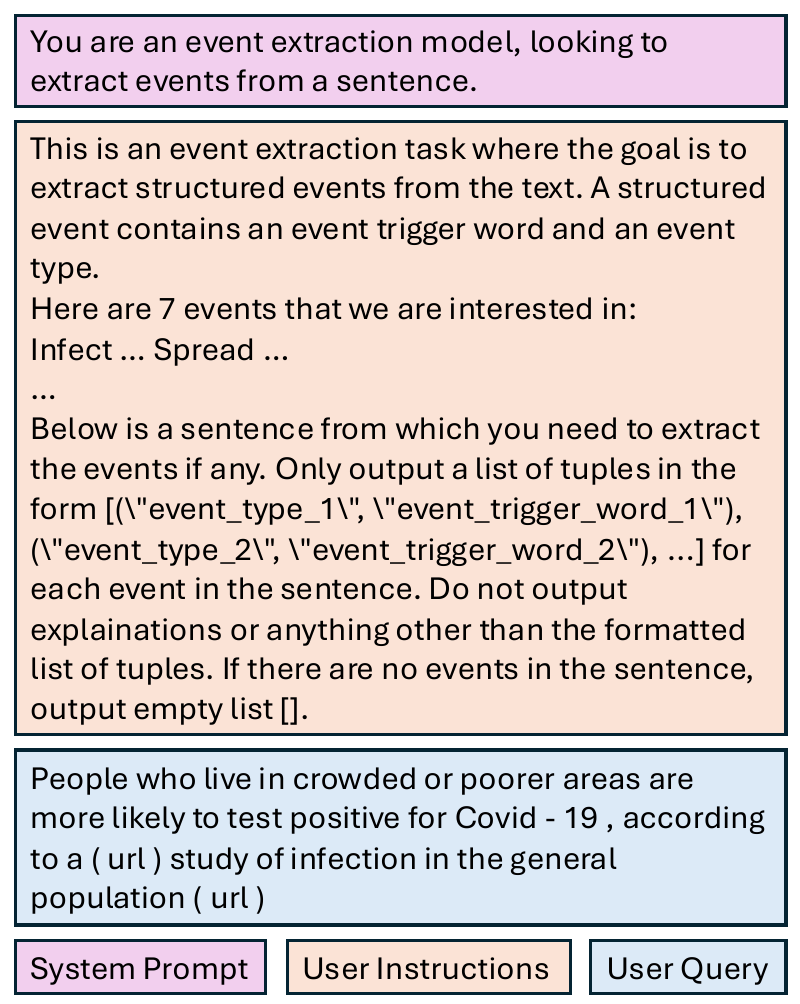}
    \caption{Illustration of the prompt utilized for multi-event direct baseline.}
    \label{fig:md-prompt}
\end{figure}

\subsection{Multi-event Direct (MD)}

Multi-event direct (MD) \cite{chatgpt-ed, huang-etal-2024-textee, llm-ee-annotator} is the most common and simplest prompting technique used for ED.
It prompts the model directly to reason across all the events and provide the relevant triggers based on the query text.
We try various prompt versions and illustrate the best engineered prompt based on a small study in Figure~\ref{fig:md-prompt}.
Majorly, we include all task-specific instructions and constraints in a single verbalized prompt, which can overload the LLM's reasoning capability.

\begin{figure}
    \centering
    \includegraphics[width=\linewidth]{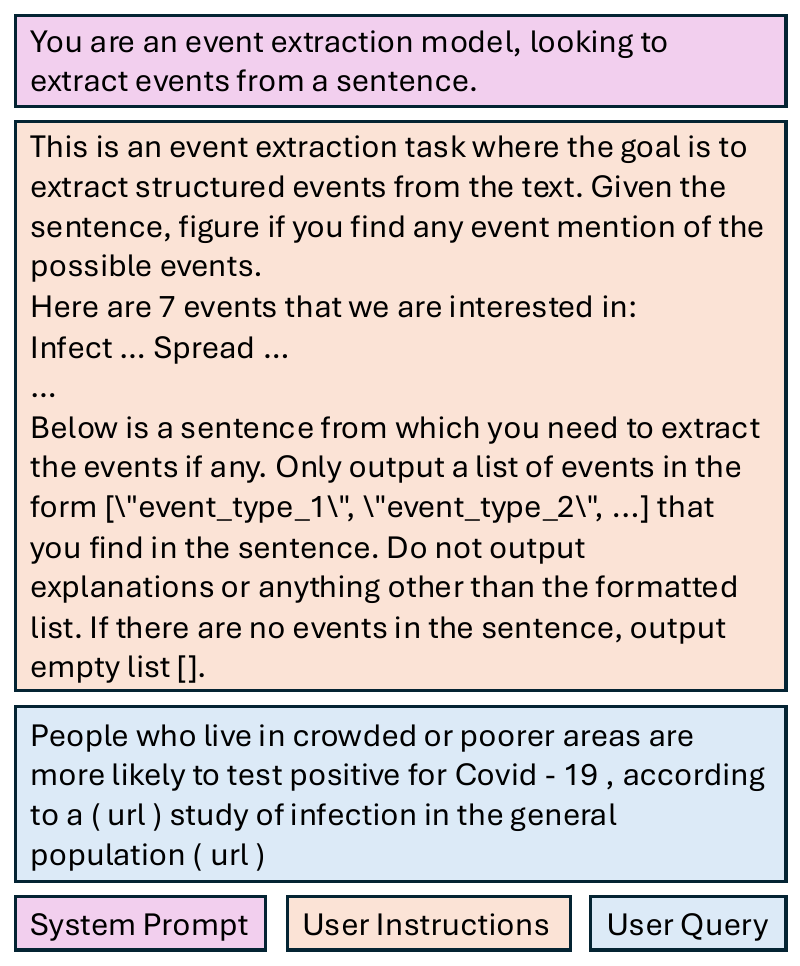}
    \caption{Illustration of the Stage-1 prompt utilized for multi-event staged baseline.}
    \label{fig:ms-stage1-prompt}
\end{figure}

\begin{figure}
    \centering
    \includegraphics[width=\linewidth]{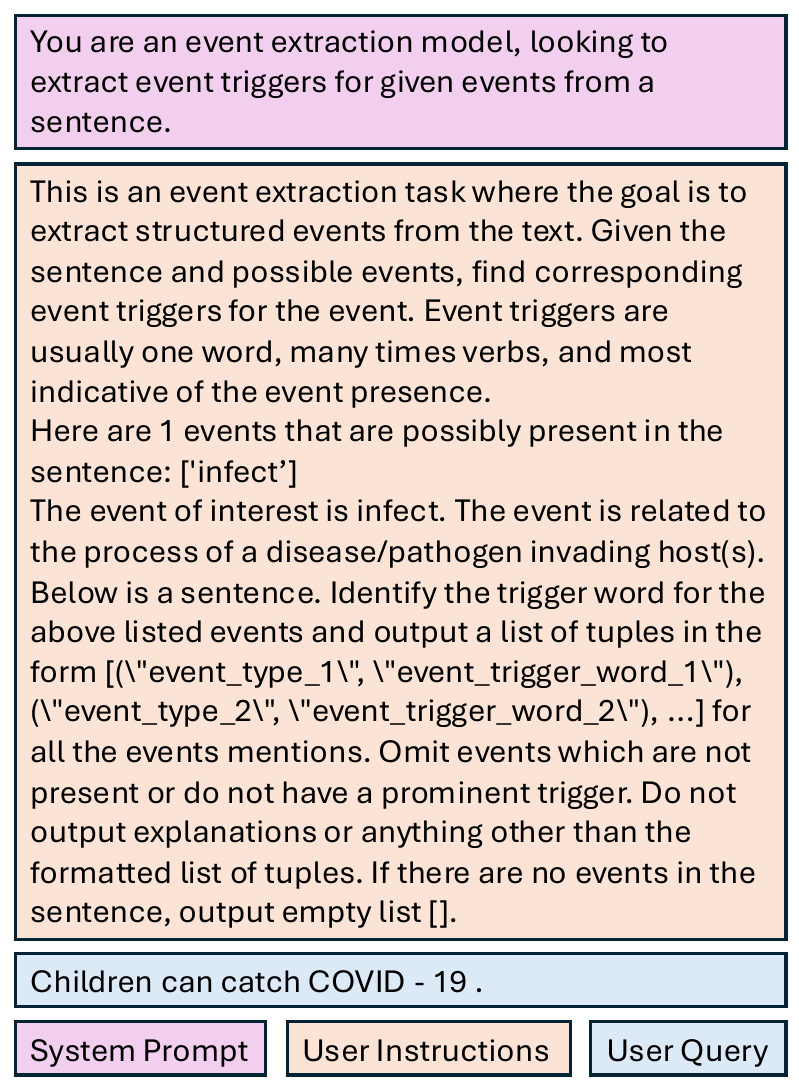}
    \caption{Illustration of the Stage-2 prompt utilized for multi-event staged baseline.}
    \label{fig:ms-stage2-prompt}
\end{figure}

\subsection{Multi-event Staged (MS)}

Multi-event staged (MS) \cite{fig} was introduced as a way of forward generation to ensure higher trigger quality.
We extend that in our work to build a strong task decomposition baseline.
Simply, this model first extracts the event types from the texts in Stage 1 and then extracts triggers specific to these event types in Stage 2.
We provide an illustration of the two stages of MS in Figures~\ref{fig:ms-stage1-prompt} and ~\ref{fig:ms-stage2-prompt}.
In this case, the first stage majorly only focuses on the event-specific constraints, while the second stage is focused on the trigger-specific ones.

\begin{figure}
    \centering
    \includegraphics[width=\linewidth]{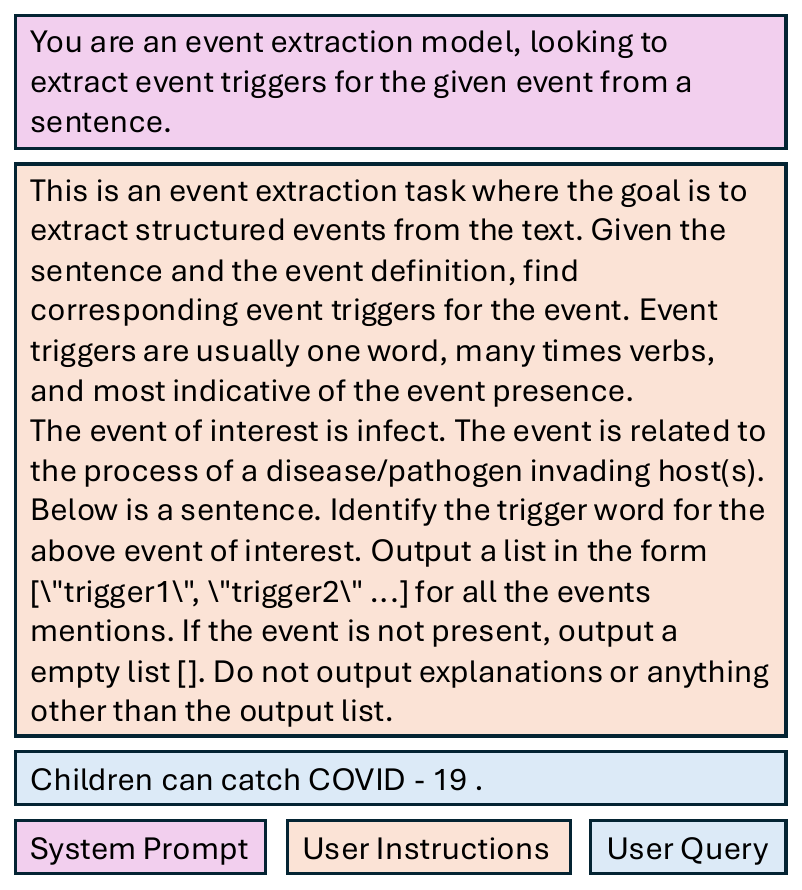}
    \caption{Illustration of the prompt utilized for binary-event direct baseline.}
    \label{fig:bd-prompt}
\end{figure}

\subsection{Binary-event Direct (BD)}

Binary-event direct (BD) \cite{lyu-etal-2021-zero, li-etal-2023-glen} has been a popular paradigm pre-dating LLMs when smaller generative text-to-text models were used.
It drastically reduces the LLM's constraints by making the LLM focus on a single event type at a time, i.e., it prompts the LLM in a multi-event direct manner, but for each event type separately.
Finally, the predictions are aggregated and output as the final prediction.
We provide an illustration of the prompt in Figure~\ref{fig:bd-prompt}.
Overall, this is a highly expensive method, especially for larger event datasets.

\begin{figure}
    \centering
    \includegraphics[width=\linewidth]{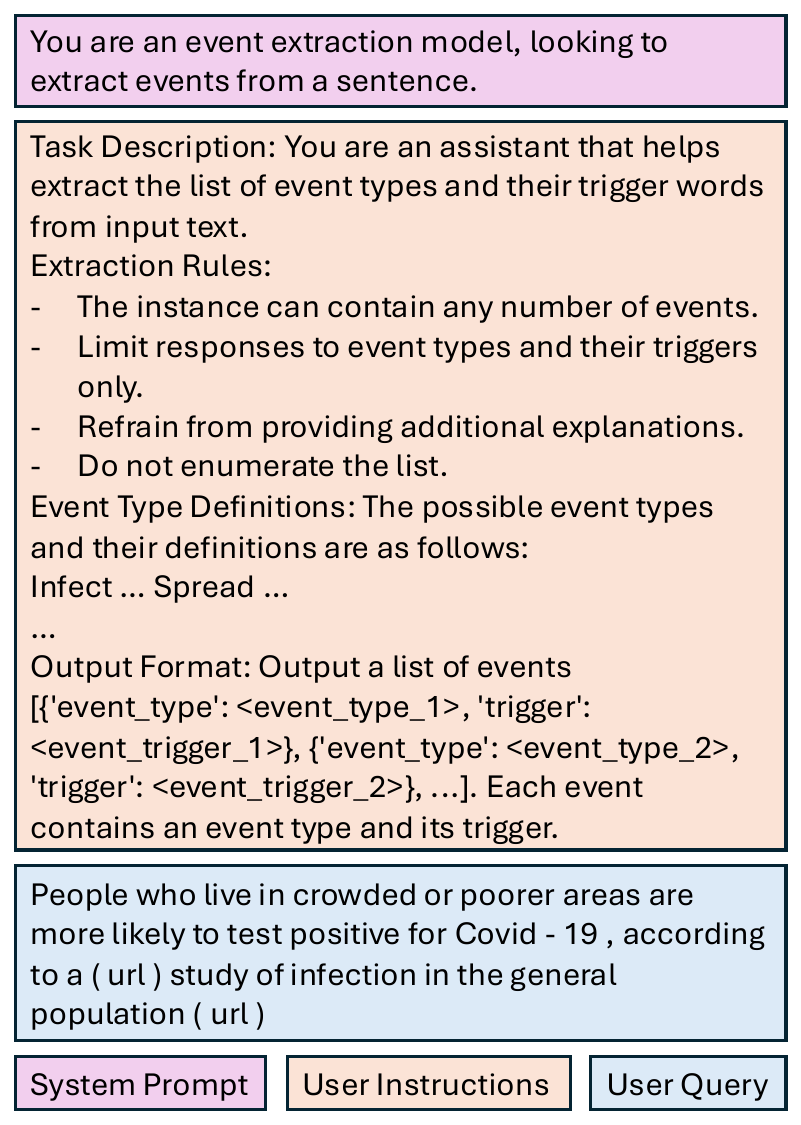}
    \caption{Illustration of the prompt utilized for Decompose-Enrich-Extract baseline.}
    \label{fig:dee-prompt}
\end{figure}

\subsection{Decompose-Enrich-Extract (DEE)}

Decompose-Enrich-Extract (DEE) \cite{dee} is a variation of the multi-event direct (MD) model, wherein it prompts the model to make predictions while enhancing the input schema.
It also puts down additional rules to make the extraction more accurate, but we posit this also adds more constraints, restricting the model's reasoning.
We provide an illustration of the prompt for this baseline in Figure~\ref{fig:dee-prompt}.

\begin{figure}
    \centering
    \includegraphics[width=\linewidth]{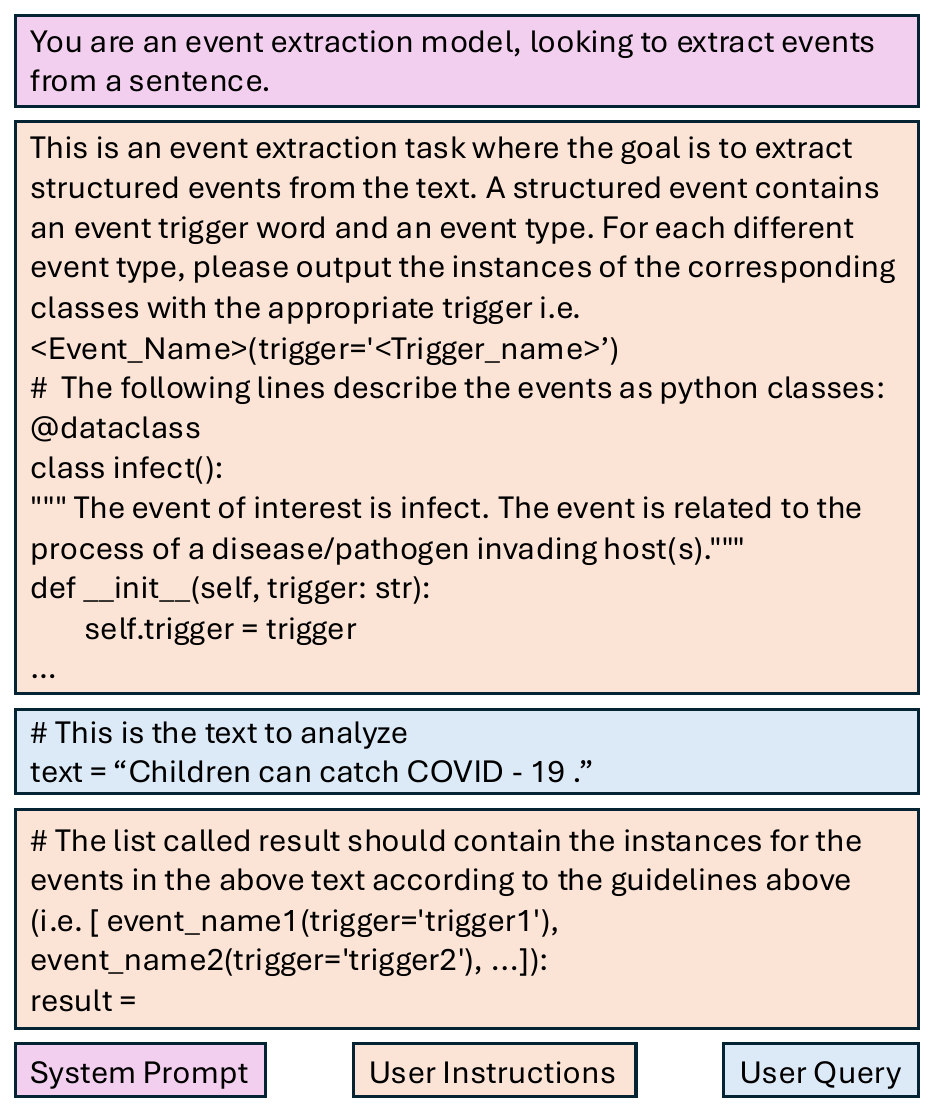}
    \caption{Illustration of the prompt utilized for GuidelineEE baseline.}
    \label{fig:gee-prompt}
\end{figure}

\subsection{GuidelineEE (GEE)}

GuidelineEE (GEE) \cite{guideline-ee} is the method focused on providing extensive guidelines to the LLM to improve its task understanding capability.
This work is similar to Code4Struct \cite{wang-etal-2023-code4struct}, wherein the input and output are more code-oriented using Python class-like structures.
The definition is provided as a docstring, and the trigger is extracted as an attribute of the class.
The output is mainly instantiations of the right set of classes.
We provide an illustration of the prompt for this baseline in Figure~\ref{fig:gee-prompt}.

\subsection{ChatIE}

ChatIE \cite{chatie} is a simple variation of multi-event staged (MS), but uses multi-turn conversation with the LLM.
Specifically, stage-1 (Figure~\ref{fig:ms-stage1-prompt}) is used as the initial prompt, and based on the output, stage-2 (Figure~\ref{fig:ms-stage2-prompt}) is used as the second turn of the prompt.

\subsection{GPT Runs}

For the GPT models (i.e., GPT3.5-turbo, GPT4o, O1-mini), we utilized Curator \cite{curator} for the API calls.
We noticed how the GPT models are already super conservative in their predictions, even when explicitly asked not to be.
The Judge component was indeed hurting model performance by making the pipeline more conservative.
Thus, we removed the Judge from all runs of the GPT LLMs.

\section{Additional Experimental Results}
\label{sec:appendix-results}

Here we provide additional and complementary results to the ones discussed in the main paper.

\subsection{Structured v/s Unstructured Output}
\label{sec:appendix-structured-output-analysis}

\begin{figure}[t]
    \centering
    \includegraphics[width=\linewidth]{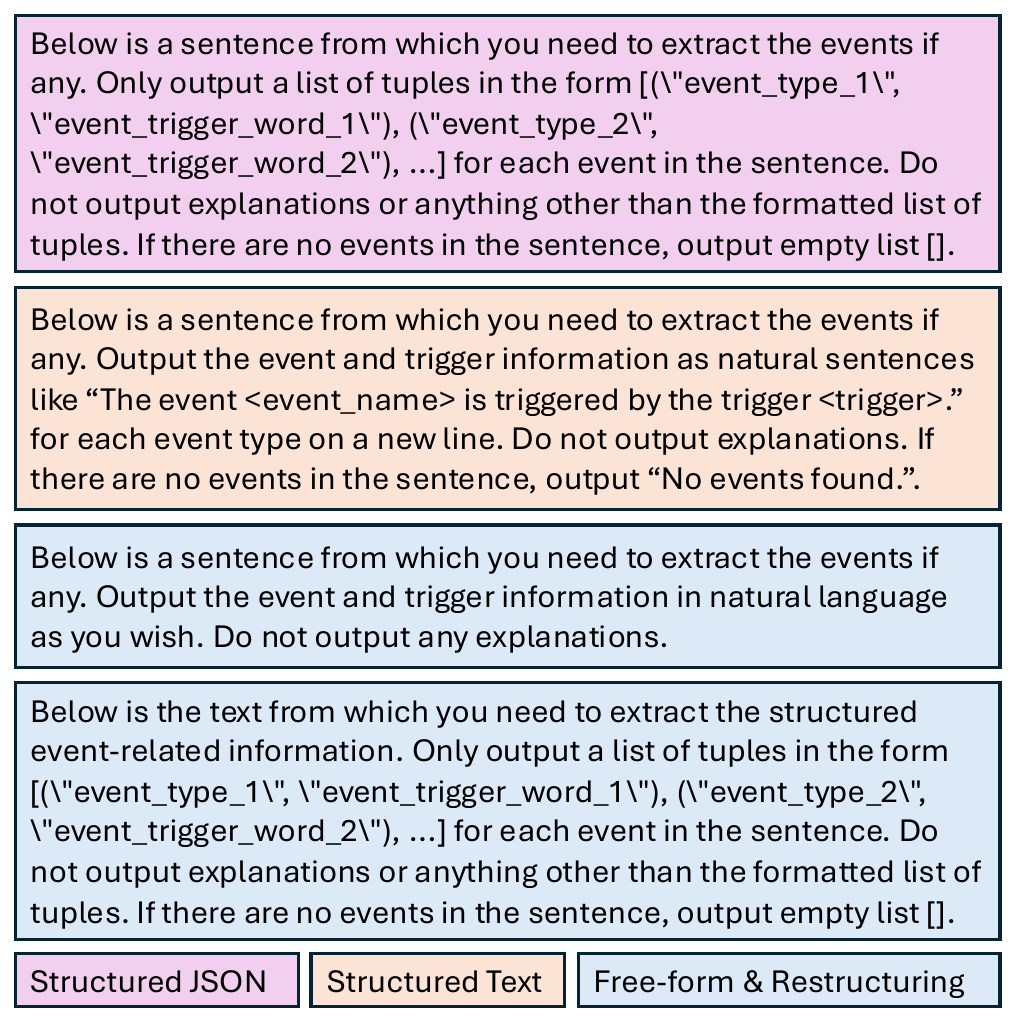}
    \caption{Illustration of the prompts utilized for the different output formats for ablating why the structured output format is better.}
    \label{fig:output-formats}
\end{figure}

In our work, we largely maintain the output to be structured to ensure easy parsing and get stronger model performance as noted in \citet{wang-etal-2023-code4struct}.
To provide more evidence, we conducted a small experiment with different output formats:
(1) Structured JSON output (the base version that we have currently) using a JSON list of tuples as the output,
(2) Structured text wherein we ask the LLM to produce natural language text but in a structured way, and
(3) Free-form text and re-structuring \cite{tam-etal-2024-speak}, wherein the LLM generates free-form text in the first generation and later restructures into JSON format using an additional LLM generation.
We provide an illustration of these output formats in Figure~\ref{fig:output-formats}.

\begin{table}[t]
    \centering
    \small
    \setlength{\tabcolsep}{3.8pt}
    \begin{tabular}{lccc}
        \toprule
        \textbf{Output Format} & \textbf{TI} & \textbf{TC} & \textbf{EI} \\
        \midrule
        Structured JSON & \textbf{35.6} & \textbf{22.4} & 30.1 \\
        Stuctured Text & 14.9 & 11.0 & \textbf{31.8} \\
        Free-form \& Restructuring & 16.7 & 12.7 & 20.8 \\
        \bottomrule
    \end{tabular}
    \caption{Ablation Study on the ACE dataset using Llama3-8B-Instruct, highlighting the significance of utilizing structured JSON output compared to text outputs.}
    \label{tab:output-format-ablation}
\end{table}

We ablate these three output formats using the Multi-event Direct (MD) prompt setting for the ACE dataset using Llama3-8B-Instruct.
We provide the results of the average of 3 runs in Table~\ref{tab:output-format-ablation}.
As clearly evidenced, any kind of text-based output format is quite poor for TI and TC metrics, highlighting the significance of JSON-based output.

\begin{table*}[t!]
\centering
\setlength{\tabcolsep}{3.5pt}
\resizebox{\textwidth}{!}{
\begin{tabular}{l|ccc|ccc|ccc}
    \toprule
    \textbf{Prompt} & \multicolumn{3}{c|}{\textbf{MAVEN} (168)} & \multicolumn{3}{c|}{\textbf{FewEvent} (100)} & \multicolumn{3}{c}{\textbf{ACE} (33)} \\
    \textbf{Style} & TI & TC & EI & TI & TC & EI & TI & TC & EI \\
    \midrule
    ChatIE
    & 33.7 ($\pm$ 0.9) & 7.3 ($\pm$ 0.6) & 13.8 ($\pm$ 0.6)
    & 20.8 ($\pm$ 0.8) & 10.2 ($\pm$ 0.6) & \textbf{27.6} ($\pm$ 0.4)
    & 30.6 ($\pm$ 1.4) & 24.9 ($\pm$ 0.9) & 46.8 ($\pm$ 0.9)
    \\
    GEE
    & 19.1 ($\pm$ 1.7) & 1.9 ($\pm$ 0.7) & 6.8 ($\pm$ 0.6)
    & 11.7 ($\pm$ 1.5) & 5.9 ($\pm$ 1.7) & 14.0 ($\pm$ 1.7)
    & 30.0 ($\pm$ 1.7) & 21.3 ($\pm$ 0.7) & 27.4 ($\pm$ 1.3)
    \\
    DEE
    & 33.7 ($\pm$ 1.4) & 6.0 ($\pm$ 0.7) & 9.2 ($\pm$ 0.4)
    & 21.1 ($\pm$ 0.5) & 10.6 ($\pm$ 0.4) & 17.8 ($\pm$ 0.2)
    & 26.9 ($\pm$ 0.8) & 19.8 ($\pm$ 0.7) & 36.1 ($\pm$ 0.8)
    \\
    BD 
    & \textbf{54.5} ($\pm$ 0.6) & 10.7 ($\pm$ 0.7) & 12.3 ($\pm$ 0.5)
    & 22.3 ($\pm$ 1.7) & 9.9 ($\pm$ 0.9) & 15.0 ($\pm$ 0.8)
    & 34.2 ($\pm$ 1.9) & 19.5 ($\pm$ 1.5) & 31.4 ($\pm$ 1.1) 
    \\
    MD
    & 45.9 ($\pm$ 1.2) & 2.8 ($\pm$ 0.2) & 4.0 ($\pm$ 0.3)
    & 25.2 ($\pm$ 0.7) & 9.5 ($\pm$ 0.2) & 15.2 ($\pm$ 0.6)
    & 35.6 ($\pm$ 1.2) & 22.4 ($\pm$ 0.8) & 30.1 ($\pm$ 0.5)
    \\
    MS
    & 46.2 ($\pm$ 1.3) & 10.3 ($\pm$ 0.7) & 11.2 ($\pm$ 0.8)
    & 20.2 ($\pm$ 1.1) & 10.2($\pm$ 0.7) & 17.0 ($\pm$ 1.1)
    & 26.7 ($\pm$ 1.4) & 17.6 ($\pm$ 0.6) & 23.1 ($\pm$ 0.9)
    \\
    \modelName
    & 53.5 ($\pm$ 1.1) & \textbf{14.4} ($\pm$ 0.7) & \textbf{17.4} ($\pm$ 0.6)
    & \textbf{26.1} ($\pm$ 0.4) & \textbf{15.7} ($\pm$ 0.7) & 25.0 ($\pm$ 0.6)
    & \textbf{40.3} ($\pm$ 1.9) & \textbf{36.3} ($\pm$ 1.2) & \textbf{47.9} ($\pm$ 0.8)
    \\
    \bottomrule
\end{tabular}}
\caption{Main results along with error bars indicating confidence intervals for the zero-shot ED performance of our proposed \modelName{} with all other baselines for the Llama3-8B-Instruct. TI: Trigger Identification, TC: Trigger Classification, EI: Event Identification. \textbf{bold} = best performance. (XX) = number of distinct event types.}
\label{tab:results-statistics-part1}
\end{table*}

\subsection{Statistical Significant Testing}
\label{sec:appendix-statistical significance}

To verify that our results are statistically significant, we provide error bars indicating confidence intervals in Table~\ref{tab:results-statistics-part1} for MAVEN, FewEvent, and ACE datasets using the Llama3-8B-Instruct.
These results demonstrate how our experimental improvements are statistically sound.
We also test and demonstrate that the improvements by \modelName{} are statistically significant (t-test using $p < 0.01$). 

\begin{table}[t]
    \centering
    \small
    \setlength{\tabcolsep}{3.8pt}
    \begin{tabular}{l|ccc|ccc}
        \toprule
        \textbf{Prompt Style} & \multicolumn{3}{c|}{\textbf{ACE}} & \multicolumn{3}{c}{\textbf{MAVEN}} \\
        & \textbf{TI} & \textbf{TC} & \textbf{EI} & \textbf{TI} & \textbf{TC} & \textbf{EI} \\
        \midrule
        MD & 36.4 & 28.6 & 34.7 & 45.9 & 3.1 & 4.4 \\
        MS & 28.2 & 20.7 & 25.2 & 45.6 & 11.1 & 12.7 \\
        \modelName{} & \textbf{47.2} & \textbf{38.3} & \textbf{48.3} & \textbf{53.2} & \textbf{15.3} & \textbf{18.7} \\
        \bottomrule
    \end{tabular}
    \caption{Ablation Study on the ACE dataset using Llama3-8B-Instruct, highlighting the significance of utilizing structured JSON output compared to text outputs.}
    \label{tab:larger-data-size}
\end{table}

\subsection{Results on larger test data size}
\label{sec:appendix-larger-data-size}

Our experimental data comprised 250 samples for evaluation to keep our findings/results consistent with the the previous work of TextEE \cite{huang-etal-2024-textee}.
Here, we provide additional experiments on larger test data size of 1000 samples for ACE and MAVEN datasets in Table~\ref{tab:larger-data-size}.
Similar to patterns in the main results, \modelName{} outperforms the baselines with gains upto 14\% F1 for ACE and 8\% F1 for MAVEN.

\subsection{Complete Results for Transfer Learning Baselines}
\label{sec:appendix-transfer-learning}

We discussed and compared \modelName{} with existing zero-shot cross-dataset transfer-learning approaches in \S~\ref{sec:transfer-learning}.
We provide complete results for each dataset in Table~\ref{tab:complete-transfer-learning-results} for a deeper analysis.
We exclude results for MAVEN and FewEvent for GOLLIE as the generations were degenerate and led to 0 F1 performance.
Across the three settings of various source-target datasets, we see how our pure zero-shot \modelName{} consistently outperforms all the fine-tuned transfer learning baselines by a considerable margin.
In fact, \modelName{}, based on the smaller Llama3-8B-Instruct LLM is stronger than most of these transfer-learning baselines.
This highlights the superior zero-shot generalization of our proposed method.

\subsection{Complete Results for Reasoning Baselines}
\label{sec:appendix-reasoning-results}

In \S~\ref{sec:reasoning-analysis}, we discuss and compare \modelName{} with reasoning-based approaches and models.
Here, we provide complete results of that comparison across datasets in Table~\ref{tab:complete-reasoning-results}.
In comparison to the non-CoT numbers, we note how CoT provides gains for the baseline models, and larger gains for the larger LLMs.
This indicates how reasoning improves model performance, but also requires more parameters and longer context handling.
Thinking-based models somehow show poorer performance compared to CoT, and our observations align with \citet{thinking-fails}.
Next, we show how the base non-CoT performance of \modelName{} is better than the CoT-based baselines.
This can also be seen when comparing thinking-based model baselines.
This strongly indicates how the strong inductive bias of \modelName{} beats the reasoning-based improvements.

Additionally, we also infuse reasoning with \modelName{}, specifically only in the Grounder stage. Reasoning in the Dreamer stage makes the model more conservative and harms the divergent reasoning we want to encourage.
We note how this additional reasoning provides further improvements of up to 1-2\% F1 over the base \modelName{} performance.

\begin{table}[t]
    \centering
    \small
    \begin{tabular}{llc}
        \toprule
        \textbf{LLM} & \textbf{Prompt Style} & \textbf{Avg. Words} \\
        \midrule
        \multirow{2}{*}{Llama3-8B} & MD + CoT & 36.8 \\
         & MS + CoT & 82.4 \\
        \multirow{2}{*}{Llama3-70B} & MD + CoT & 87.4 \\
         & MS + CoT & 107.9 \\
        \multirow{2}{*}{Qwen2.5-72B} & MD + CoT & 96.3 \\
         & MS + CoT & 184.4 \\
        \multirow{2}{*}{DS-Qwen-32B} & MD & 247.8 \\
         & MS & 525.5 \\
        \multirow{2}{*}{DS-L3-70B} & MD & 258.9 \\
         & MS & 484.4 \\
        \midrule
        Llama3-8B & \modelName & 11.6 \\
        Llama3-70B & \modelName & 6.6 \\
        Qwen2.5-72B & \modelName & 5.1 \\
        \bottomrule
    \end{tabular}
    \caption{Efficiency analysis in terms of average number of words per query (Avg. Words) of \modelName{} with other reasoning-based baselines on the ACE dataset.}
    \label{tab:efficiency-reasoning}
\end{table}

\begin{table}[t]
    \centering
    \small
    \begin{tabular}{l|ccc}
        \toprule
        \textbf{Component/LLM} & \multicolumn{3}{c}{\textbf{EI}} \\
        & P & R & F \\
        \midrule
        \multicolumn{4}{c}{\textbf{Llama3-8B-Instruct}} \\
        \midrule
        Dreamer & 0.0 & 0.0 & 0.0 \\
        \quad + Grounder & 19.1 & 45.6 & 26.9 \\
        \quad + FSM Decoding & 21.1 & 54.9 & 32.3 \\
        \quad + Judge & \textbf{49.5} & \textbf{46.5} & \textbf{47.9} \\
        \midrule
        MD Baseline & 40.5 & 23.9 & 30.1 \\
        MS Baseline & 18.9 & 29.6 & 23.1 \\
        \midrule
        \multicolumn{4}{c}{\textbf{Llama3-70B-Instruct}} \\
        \midrule
        Dreamer & 0.0 & 0.0 & 0.0 \\
        \quad + Grounder & 25.4 & 61.5 & 36.0 \\
        \quad + FSM Decoding & 29.1 & 60.1 & 39.2 \\
        \quad + Judge & 50.8 & \textbf{60.1} & \textbf{55.1} \\
        \midrule
        MD Baseline & 51.2 & 43.2 & 46.8 \\
        MS Baseline & \textbf{62.5} & 37.5 & 46.9 \\
        \bottomrule
    \end{tabular}
    \caption{Ablation Study using Trigger Identification (TI) on the ACE dataset highlighting the significance and contribution of each component of \modelName. P: Precision, R: Recall, F: F1 score.}
    \label{tab:appendix-ablation-study}
\end{table}

\begin{table*}[t!]
\centering
\setlength{\tabcolsep}{3.5pt}
\resizebox{\textwidth}{!}{
\begin{tabular}{ll|ccc|ccc|ccc|ccc|ccc|ccc|ccc}
    \toprule
    \multirow{2}{*}{\textbf{LM/LLM}} & \textbf{Prompt} & \multicolumn{3}{c|}{\textbf{MAVEN} (168)} & \multicolumn{3}{c|}{\textbf{FewEvent} (100)} & \multicolumn{3}{c|}{\textbf{ACE} (33)} & \multicolumn{3}{c|}{\textbf{GENIA} (9)} & \multicolumn{3}{c|}{\textbf{SPEED} (7)} & \multicolumn{3}{c|}{\textbf{CASIE} (5)} & \multicolumn{3}{c}{\textbf{Average}} \\
    & \textbf{Style} & TI & TC & EI & TI & TC & EI & TI & TC & EI & TI & TC & EI & TI & TC & EI & TI & TC & EI & TI & TC & EI \\
    \midrule
    \multicolumn{23}{c}{\textbf{Trained on ACE data* \quad $\rightarrow$ \quad Tested on other datasets}} \\
    \midrule
    BART-large & DEGREE 
    & 29.4 & 11.0 & 13.8 
    & 42.6 & 22.5 & 27.2
    & - & - & -
    & 5.1 & 3.5 & 11.6
    & 23.4 & 16.2 & 26.7
    & 3.8 & 2.0 & 27.0
    & 20.9 & 11.0 & 21.3 \\
    Llama3-8B & \modelName
    & 53.5 & 14.4 & 17.4 
    & 26.1 & 15.7 & 25.0 
    & - & - & - 
    & 25.8 & 15.4 & 30.0 
    & 35.5 & 23.6 & 42.4 
    & 18.5 & 16.8 & 58.8
    & 31.9 & 17.2 & 34.7 \\
    Llama3-70B & \modelName
    & 62.5 & 27.8 & 30.6
    & 40.4 & 25.1 & 36.1 
    & - & - & -
    & 38.6 & 31.0 & 48.5
    & 45.0 & 36.5 & 51.8
    & 17.3 & 16.6 & 66.6
    & 40.8 & 27.4 & 46.7 \\
    GPT4o & \modelName
    & 58.5 & 32.2 & 35.6 
    & 36.1 & 28.4 & 38.5
    & - & - & -
    & 40.7 & 35.4 & 51.2 
    & 43.3 & 37.3 & 46.1 
    & 16.7 & 16.7 & 58.8 
    & 39.1 & 30.0 & 46.0 \\
    \midrule
    \multicolumn{23}{c}{\textbf{Trained on MAVEN data* \quad $\rightarrow$ \quad Tested on other datasets}} \\
    \midrule
    BART-large & DEGREE 
    & - & - & -
    & 31.1 & 18.7 & 25.0
    & 43.3 & 36.6 & 38.2
    & 33.9 & 27.6 & 46.2
    & 44.8 & 37.1 & 44.8
    & 6.1 & 5.2 & 38.6
    & 31.8 & 25.0 & 38.6 \\
    Llama3-8B & \modelName
    & - & - & -
    & 26.1 & 15.7 & 25.0 
    & 40.3 & 36.3 & 47.9 
    & 25.8 & 15.4 & 30.0 
    & 35.5 & 23.6 & 42.4 
    & 18.5 & 16.8 & 58.8
    & 29.2 & 21.6 & 40.8 \\
    Llama3-70B & \modelName
    & - & - & -
    & 40.4 & 25.1 & 36.1 
    & 57.2 & 49.5 & 55.1 
    & 38.6 & 31.0 & 48.5
    & 45.0 & 36.5 & 51.8
    & 17.3 & 16.6 & 66.6
    & 39.7 & 31.7 & 51.6 \\
    GPT4o & \modelName
    & - & - & -
    & 36.1 & 28.4 & 38.5
    & 54.9 & 54.9 & 56.6 
    & 40.7 & 35.4 & 51.2 
    & 43.3 & 37.3 & 46.1 
    & 16.7 & 16.7 & 58.8 
    & 38.3 & 34.5 & 50.2 \\
    \midrule
    \multicolumn{23}{c}{\textbf{Trained on ACE data* \quad $\rightarrow$ \quad Tested on GENIA, SPEED, CASIE}} \\
    \midrule
    GOLLIE-7B & GOLLIE 
    & - & - & - 
    & - & - & -
    & - & - & -
    & 3.2 & 2.2 & 7.1
    & 12.6 & 11.6 & 24.3
    & 2.1 & 2.1 & 14.4
    & 6.0 & 5.3 & 15.3 \\
    GOLLIE-34B & GOLLIE
    & - & - & - 
    & - & - & -
    & - & - & -
    & 26.5 & 22.8 & 40.4
    & 15.9 & 10.9 & 19.1
    & 4.5 & 1.5 & 28.6
    & 15.6 & 11.7 & 29.4 \\
    Llama3-8B & \modelName
    & - & - & - 
    & - & - & - 
    & - & - & - 
    & 25.8 & 15.4 & 30.0 
    & 35.5 & 23.6 & 42.4 
    & 18.5 & 16.8 & 58.8
    & 26.6 & 18.6 & 43.7 \\
    Llama3-70B & \modelName
    & - & - & - 
    & - & - & - 
    & - & - & -
    & 38.6 & 31.0 & 48.5
    & 45.0 & 36.5 & 51.8
    & 17.3 & 16.6 & 66.6
    & 33.6 & 28.0 & 55.6 \\
    GPT4o & \modelName
    & - & - & - 
    & - & - & - 
    & - & - & -
    & 40.7 & 35.4 & 51.2 
    & 43.3 & 37.3 & 46.1 
    & 16.7 & 16.7 & 58.8 
    & 33.6 & 29.8 & 52.0 \\
    \bottomrule
\end{tabular}}
\caption{Complete results for comparison of \modelName{} with other fine-tuned transfer-learning approaches for zero-shot ED. *Training done for models other than \modelName. \modelName{} results are pure zero-shot, i.e., without any training. "-" indicates training data or where results were degenerate. (XX) = number of distinct event types.}
\label{tab:complete-transfer-learning-results}
\end{table*}

\begin{table}[t]
    \centering
    \small
    \begin{tabular}{p{3.8cm}|p{3cm}}
        \toprule
        \textbf{Sentence} & \textbf{Baseline Prediction} \\
        \midrule
        \multicolumn{2}{c}{\textbf{Precision Errors}} \\
        \midrule
        In the near future we will be expanding this to include all the other organizations that we can contact, but we are just keeping things safe for now. & [(\redtext{"Phone-Write", "contact"})] \\
        \hline
        The Holocaust of the Jews and Zigeuner was motivated by racial prejudices. & [(\redtext{"Attack"}, \greentext{"Holocaust"})]\\
        \hline
        My friend, an ER physician has said over 70\% of people who test positive for covid NEVER have a fever. & [(\redtext{"symptom", "fever"})] \\
        \hline
        On 4 April 2013, a building collapsed on tribal land in Mumbra. & [(\redtext{"Destroying"}, \greentext{"collapsed"})]\\
        \midrule
        \multicolumn{2}{c}{\textbf{Recall Errors}} \\
        \midrule
        Pasko was released in January for good behavior after serving more than two-thirds of the sentence. &  [(\greentext{"Release-Parole", "released"})]\newline \redtext{Missed: ("Sentence", "sentence")} \\
        \hline
        People who live in crowded or poorer areas are more likely to test positive for Covid - 19 & [] \newline \redtext{Missed: ("infect", "positive")} \\
        \hline
        WOW debuted on January 18 as part of AXS's Friday Night Fights schedule & [] \newline \redtext{Missed: ("Process\_start", "debuted")} \\
        \hline
        He is got it pretty easy Id say even with the international travel & [] \newline \redtext{Missed: ("Transport-person", "travel")} \\
        \bottomrule
    \end{tabular}
    \caption{Qualitative examples highlighting the various errors by zero-shot LLM baselines. We highlight the correct predictions in \greentext{green} and incorrect ones in \redtext{red}.}
    \label{tab:baseline-errors}
\end{table}

\begin{table*}[t!]
\centering
\setlength{\tabcolsep}{3pt}
\resizebox{\textwidth}{!}{
\begin{tabular}{ll|ccc|ccc|ccc|ccc|ccc|ccc|ccc}
    \toprule
    \multirow{2}{*}{\textbf{LLM}} & \textbf{Prompt} & \multicolumn{3}{c|}{\textbf{MAVEN} (168)} & \multicolumn{3}{c|}{\textbf{FewEvent} (100)} & \multicolumn{3}{c|}{\textbf{ACE} (33)} & \multicolumn{3}{c|}{\textbf{GENIA} (9)} & \multicolumn{3}{c|}{\textbf{SPEED} (7)} & \multicolumn{3}{c|}{\textbf{CASIE} (5)} & \multicolumn{3}{c}{\textbf{Average}} \\
    & \textbf{Style} & TI & TC & EI & TI & TC & EI & TI & TC & EI & TI & TC & EI & TI & TC & EI & TI & TC & EI & TI & TC & EI \\
    \midrule
    \multicolumn{23}{c}{\textbf{Chain-of-thought}} \\
    \midrule
    \multirow{6}{*}{Llama3-8B} & MD
    & 45.9 & 2.8 & 4.0
    & 25.2 & 9.5 & 15.2
    & 35.6 & 22.4 & 30.1
    & 22.8 & 15.3 & 25.4
    & 34.9 & 27.8 & 42.4
    & 10.3 & 8.8 & 47.9
    & 29.1 & 14.4 & 27.5 
    \\
    & \quad + CoT
    & 35.4 & 3.2 & 4.8
    & 15.4 & 6.8 & 13.8
    & 30.6 & 18.7 & 27.6
    & 24.3 & 15.9 & 26.9
    & 34.6 & 27.8 & 42.1
    & 9.8 & 8.7 & 47.1
    & 25.0 & 13.5 & 27.1
    \\
    & MS
    & 46.2 & 10.3 & 11.2
    & 20.2 & 10.2 & 17.0 
    & 26.7 & 17.6 & 23.1 
    & \textbf{27.6} & 19.7 & 30.5 
    & 34.1 & 27.3 & 40.6
    & 11.9 & 10.3 & 48.3 
    & 27.8 & 15.9 & 28.4
    \\
    & \quad + CoT
    & 35.9 & 7.2 & 8.2
    & 20.5 & 11.1 & 19.3
    & 34.3 & 23.4 & 32.9
    & 27.2 & \textbf{20.1} & 29.6
    & \textbf{39.4} & \textbf{31.9} & \textbf{46.6}
    & 13.1 & 12.2 & 54.8
    & 28.4 & 17.6 & 31.9
    \\
    & \modelName
    & 53.5 & 14.4 & 17.4 
    & 26.1 & \textbf{15.7} & \textbf{25.0} 
    & \textbf{40.3} & 36.3 & \textbf{47.9} 
    & 25.8 & 15.4 & 30.0 
    & 35.5 & 23.6 & 42.4 
    & \textbf{18.5} & \textbf{16.8} & \textbf{58.8}
    & \textbf{33.3} & 20.4 & \textbf{36.9}
    \\
    & \quad + CoT
    & \textbf{53.6} & \textbf{15.5} & \textbf{17.9}
    & \textbf{27.5} & 15.4 & 24.7
    & 39.8 & \textbf{36.6} & 45.0
    & 25.8 & 16.4 & \textbf{31.9}
    & 35.1 & 26.6 & 41.5
    & 16.7 & 15.9 & 56.0
    & 33.1 & \textbf{21.1} & 36.2
    \\
    \midrule
    \multirow{6}{*}{Llama3-70B} & MD
    & \textbf{63.5} & 14.2 & 14.7
    & 34.0 & 20.9 & 32.6
    & 51.2 & 40.2 & 46.8
    & 36.8 & 28.9 & 43.0
    & 45.4 & 36.8 & 49.0
    & 13.9 & 13.7 & 64.4 
    & 40.8 & 25.8 & 41.8 
    \\
    & \quad + CoT
    & 56.0 & 29.4 & 32.5
    & 37.1 & 25.3 & 37.2
    & 54.9 & 48.5 & 57.1
    & 35.4 & 28.2 & 45.5
    & 47.1 & 39.5 & 50.3
    & 15.7 & 14.8 & 65.4
    & 41.0 & 30.9 & 48.0
    \\
    & MS
    & 33.9 & 21.6 & 22.3 
    & 35.3 & 24.9 & \textbf{39.9}
    & 49.9 & 42.8 & 46.9
    & 37.4 & 31.0 & 45.0
    & 43.8 & 35.5 & 49.6 
    & 14.0 & 14.0 & 59.5
    & 35.7 & 28.3 & 43.9 
    \\
    & \quad + CoT
    & 55.7 & 29.5 & 32.6
    & 34.9 & 25.4 & 38.6
    & 56.1 & 51.3 & 56.5
    & 31.8 & 26.4 & 37.7
    & \textbf{49.7} & \textbf{42.5} & \textbf{56.6}
    & 14.8 & 14.6 & 60.6
    & 40.5 & 31.6 & 47.1 
    \\
    & \modelName
    & 62.5 & 27.8 & 30.6
    & 40.4 & 25.1 & 36.1 
    & \textbf{57.2} & 49.5 & 55.1 
    & \textbf{38.6} & 31.0 & \textbf{48.5}
    & 45.0 & 36.5 & 51.8
    & 17.3 & 16.6 & 66.6
    & \textbf{43.5} & 32.8 & 48.1
    \\
    & \quad + CoT
    & 61.2 & \textbf{34.1} & \textbf{36.4}
    & \textbf{40.9} & \textbf{27.3} & 37.5
    & 55.4 & \textbf{51.7} & \textbf{58.5}
    & 37.9 & \textbf{31.7} & 48.1
    & 44.3 & 36.5 & 50.8
    & \textbf{18.0} & \textbf{17.4} & \textbf{67.1}
    & 43.0 & \textbf{33.1} & \textbf{49.8} \\
    \midrule
    \multirow{6}{*}{Qwen2.5-72B} & MD
    & 49.4 & 21.6 & 24.1 
    & 17.0 & 12.3 & 21.0 
    & 28.8 & 25.8 & 30.3 
    & 30.5 & 27.0 & 36.3
    & 41.4 & 37.4 & 45.4
    & 11.0 & 10.4 & 57.9
    & 29.7 & 22.4 & 35.8 
    \\
    & \quad + CoT
    & 54.0 & 27.9 & \textbf{33.8}
    & 26.7 & 20.5 & 33.3
    & 46.1 & 41.6 & 47.3
    & 29.5 & 26.1 & 38.9
    & \textbf{42.6} & 36.8 & \textbf{48.1}
    & 10.3 & 9.9 & \textbf{60.0}
    & 34.9 & 27.1 & 43.6
    \\
    & MS
    & 39.9 & 23.6 & 25.4
    & 25.0 & 21.0 & 34.2
    & 42.5 & 40.4 & 42.5
    & 26.7 & 23.6 & 34.1
    & 40.6 & 35.5 & 45.2
    & 10.5 & 10.5 & 49.1
    & 30.9 & 25.8 & 38.4 
    \\
    & \quad + CoT
    & \textbf{54.2} & 28.0 & 31.1
    & 28.3 & 21.5 & 33.6
    & \textbf{48.5} & \textbf{46.3} & \textbf{48.9}
    & 30.7 & 26.5 & 38.7
    & 44.9 & \textbf{39.7} & 47.9
    & 10.6 & 10.6 & 44.5
    & 36.2 & 28.8 & 40.8
    \\
    & \modelName
    & 54.1 & 27.5 & 30.2
    & 30.8 & 22.3 & 32.9
    & 46.8 & 44.8 & 47.8
    & 33.6 & \textbf{29.8} & \textbf{43.9}
    & 40.6 & 34.7 & 41.4
    & 15.9 & 15.8 & 59.3
    & 37.0 & 29.2 & 42.6
    \\
    & \quad + CoT
    & \textbf{54.2} & \textbf{29.7} & \textbf{33.8}
    & \textbf{31.7} & \textbf{23.5} & \textbf{35.5}
    & 45.4 & 42.2 & 45.4
    & \textbf{34.2} & 29.2 & 43.6
    & 40.5 & 34.6 & 44.8
    & \textbf{16.8} & \textbf{16.7} & \textbf{60.0}
    & \textbf{37.1} & \textbf{29.3} & \textbf{43.8}
    \\
    \midrule
    \multicolumn{23}{c}{\textbf{Thinking-based models}} \\
    \midrule
    \multirow{3}{*}{DS-Qwen-32B} & MD 
    & 55.3 & 26.7 & 30.1
    & 34.0 & 23.7 & 36.8
    & \textbf{56.3} & 51.8 & 60.2
    & 33.2 & 27.5 & 41.2
    & 45.5 & 39.0 & 54.5
    & 11.1 & 11.1 & 54.9
    & 39.2 & 30.0 & 46.3
    \\
    & MS 
    & 55.0 & 25.8 & 29.6
    & 33.8 & 23.3 & \textbf{38.5}
    & 50.6 & 48.9 & 59.6
    & 30.5 & 25.0 & 36.6
    & \textbf{52.7} & \textbf{44.7} & 54.7
    & 14.6 & 14.6 & 51.9
    & 39.5 & 30.4 & 45.2
    \\
    & \modelName 
    & \textbf{60.1} & \textbf{30.2} & \textbf{32.6}
    & \textbf{38.5} & \textbf{26.1} & 36.8
    & \textbf{56.3} & \textbf{53.9} & \textbf{60.5}
    & \textbf{36.3} & \textbf{30.4} & \textbf{47.6}
    & 48.6 & 41.1 & \textbf{55.2}
    & \textbf{18.5} & \textbf{17.8} & \textbf{64.4}
    & \textbf{43.1} & \textbf{33.3} & \textbf{49.5}
    \\
    \midrule
    \multirow{3}{*}{DS-L3-70B} & MD 
    & 48.3 & 31.2 & 32.5
    & 13.7 & 9.6 & 17.3
    & 31.5 & 27.8 & 34.5
    & 24.5 & 21.6 & 31.9
    & 45.3 & 38.9 & 50.6
    & 10.5 & 10.5 & 50.0
    & 29.0 & 23.3 & 36.1
    \\
    & MS 
    & 50.3 & 28.3 & 31.3
    & 23.9 & 18.5 & 28.3
    & 36.8 & 33.7 & 38.0
    & 27.8 & 24.6 & 35.3
    & 48.2 & \textbf{44.2} & 49.2
    & 12.6 & 12.6 & 44.7
    & 33.3 & 27.0 & 37.8
    \\
    & \modelName 
    & \textbf{59.5} & \textbf{34.7} & \textbf{37.2}
    & \textbf{36.2} & \textbf{25.9} & \textbf{35.0}
    & \textbf{53.0} & \textbf{51.3} & \textbf{55.8}
    & \textbf{32.3} & \textbf{28.6} & \textbf{42.7}
    & \textbf{49.3} & 39.8 & \textbf{53.4}
    & \textbf{18.0} & \textbf{17.9} & \textbf{65.9}
    & \textbf{41.4} & \textbf{33.0} & \textbf{48.3}
    \\
    \midrule
    O1-mini & MD 
    & 59.1 & 32.8 & 35.7
    & 36.8 & 28.0 & 40.3
    & 53.9 & 48.5 & 53.0
    & 35.8 & 33.7 & 43.8 
    & 44.2 & 40.2 & 48.1 
    & 11.5 & 11.5 & 47.5 
    & 40.2 & 32.5 & 44.7 \\
    \bottomrule
\end{tabular}}
\caption{Complete results for comparison of \modelName{} with reasoning approaches like Chain-of-thought (CoT) and thinking-based models for zero-shot ED. \textbf{bold} = best performance. (XX) = number of distinct event types.}
\label{tab:complete-reasoning-results}
\end{table*}

\paragraph{Efficiency analysis:}
Apart from performance, we also analyze the effectiveness in terms of efficiency of the various methods.
We measure efficiency by the average number of output words generated per query (which should be equivalent to the average number of output tokens).
We provide this comparison for the different methods and LLMs for the ACE dataset in Table~\ref{tab:efficiency-reasoning}.
As evident, CoT and thinking-based models expend a large amount of tokens on token-based reasoning, which is zero in the case of \modelName.
On average, \modelName{} reduces the output words by 15x compared to CoT and by up to 55x compared to the thinking-based models.
This highlights the practical utility of \modelName{} where it can provide higher performance at vastly reduced token generation cost.

\subsection{Additional results for Ablation Study}
\label{sec:appendix-ablation-study}

We provided an ablation study for \modelName's components in \S~\ref{sec:ablation-study}.
Here we provide additional results for the same study, specifically for the Event Identification (EI) evaluation metric in Table~\ref{tab:appendix-ablation-study}.
We conclude observations similar to those noted in the main paper, highlighting how \modelName{} helps increase the recall without much decreasing the precision of the model.
Dreamer has a 0\% score since the event names are free-form text generations in this stage.

\begin{table*}[t]
    \centering
    \small
    \begin{tabular}{p{4cm}|p{3.6cm}|p{3.6cm}|p{3cm}}
        \toprule
        \textbf{Sentence} & \textbf{Dreamer } & \textbf{Grounder} & \textbf{Judge} \\
        & \textbf{Prediction} & \textbf{Prediction} & \textbf{Prediction} \\
        \midrule
        Police also arrested two Moroccan men suspected of trafficking in human beings and navigating the Zodiac boat across from Africa, Efe said. & [("arrest", "arrested"), ("trafficking", "trafficking"), ("navigating", "navigating"), ("said", "said")] & [("Arrest-Jail", "arrested"), ("Charge-Indict", "trafficking")] & [(\greentext{"Arrest-Jail", "arrested"})]\\
        \midrule
        Only 4 men have competed without eliminating a single opponent Fire, Mini Maximo, Sombrita and Stukita. & [("compete", "competed"), ("eliminate", "eliminating")] & [("Competition", "competed")] & [(\greentext{"Competition", "competed")}]\\
        \midrule
        Weird as hell: the Covid-19 patients who have symptoms for months | Coronavirus outbreak | The Guardian (url) & [("Disease\_Spread", "outbreak"), ("Infection", "patients"), ("Symptom\_Show", "symptoms")] & [("symptom", "symptoms"), ("spread", "outbreak")] & [(\greentext{"symptom", "symptoms"}), (\redtext{"spread", "outbreak"})] \\
        \midrule
         The time he has spent inside roughly equates to 2 years per woman he killed & [("Kill", "killed"), ("Spend", "spent"), ("Equate", "equates")] & [("Life.Die", "killed")] & [(\greentext{"Life.Die", "killed"})] \\
        \bottomrule
    \end{tabular}
    \caption{Qualitative examples eliciting \modelName's predictions per component for various input sentences. We highlight the correct predictions in \greentext{green} and incorrect ones in \redtext{red}.}
    \label{tab:dicore-preds}
\end{table*}

\section{Broader Qualitative Study}
\label{sec:appendix-qual-study}

We provided a brief qualitative study eliciting some common errors of previous baselines and how \modelName fixes them in \S~\ref{sec:ablation-study}.
Here, we provide some more examples to highlight the various errors made by previous baselines in Table~\ref{tab:baseline-errors}.
Next, we also show some more examples to elicit the internal component-wise predictions of \modelName{} in Table~\ref{tab:dicore-preds}.
Overall, these examples demonstrate the utility of the divergent-convergent reasoning paradigm for ED.

\end{document}